%% file: aaai24.tex
\documentclass[letterpaper]{article} 
\usepackage{aaai24}  
\usepackage{times}  
\usepackage{helvet}  
\usepackage{courier}  
\usepackage[hyphens]{url}  
\usepackage{graphicx} 
\usepackage{natbib}  
\usepackage{caption} 
\usepackage{algorithm}
\usepackage{algorithmic}

\usepackage{subcaption}
\usepackage[dvipsnames]{xcolor}
\usepackage{microtype}      

\usepackage{newfloat}
\usepackage{listings}
\DeclareCaptionStyle{ruled}{labelfont=normalfont,labelsep=colon,strut=off} 
\floatstyle{ruled}
\newfloat{listing}{tb}{lst}{}
\floatname{listing}{Listing}
\title{Working Memory Capacity of ChatGPT:
An Empirical Study}
\author {
    Dongyu Gong\textsuperscript{\rm 1,2},
    Xingchen Wan\textsuperscript{\rm 1},
    Dingmin Wang\textsuperscript{\rm 1}
}
\affiliations {
    \textsuperscript{\rm 1}University of Oxford\\
    \textsuperscript{\rm 2}Yale University\\
    dongyu.gong@yale.edu, xwan@robots.ox.ac.uk, dingmin.wang@cs.ox.ac.uk
}

\usepackage{bibentry}

\begin{document}

\maketitle

\input{1-abstract}
\input{2-introduction}

\input{3-background}

\input{4-methodology}

\input{5-experiments}

\input{6-conclusion}
\bibliography{aaai24}
\onecolumn
\input{7-supplement}
\end{document}

%% file: 1-abstract.tex
\begin{abstract}
Working memory is a critical aspect of both human intelligence and artificial intelligence, serving as a workspace for the temporary storage and manipulation of information. In this paper, we systematically assess the working memory capacity of ChatGPT, a large language model developed by OpenAI, by examining its performance in verbal and spatial \textit{n}-back tasks under various conditions. Our experiments reveal that ChatGPT has a working memory capacity limit strikingly similar to that of humans. Furthermore, we investigate the impact of different instruction strategies on ChatGPT's performance and observe that the fundamental patterns of a capacity limit persist. From our empirical findings, we propose that \textit{n}-back tasks may serve as tools for benchmarking the working memory capacity of large language models and hold potential for informing future efforts aimed at enhancing AI working memory.
\end{abstract}

%% file: 2-introduction.tex
\section{Introduction}
The advent of large language models (LLMs) like ChatGPT and GPT-4 \cite{OpenAI2023GPT4TR} has propelled the pursuit of artificial general intelligence~\cite{bubeck2023sparks} and unveiled human-level emergent abilities~\cite{wei2022emergent, kosinski2023theory}. Among these abilities is the capacity to retain contextual information while engaging in multi-turn conversations, suggesting the presence of working memory in these LLMs.

In cognitive science, working memory is usually defined as the ability to store and manipulate information in mind~\cite{baddeley1992working} temporarily. It is widely regarded as a critical element of human intelligence, as it underlies various higher-order cognitive processes such as reasoning, problem-solving, and language comprehension~\cite{conway_kovacs_2020}. 

Studies on human participants have revealed a fundamental capacity limit in working memory~\cite{cowan2001magical} but there is no consensus on why working memory capacity is limited~\cite{oberauerWhatLimitsWorking2016,wilhelmWhatWorkingMemory2013}. Among many theories, the executive attention hypothesis~\cite{engle_kane_tuholski_1999,engleWorkingMemoryCapacity2002} suggests that working memory depends on utilizing attention to maintain or suppress information, and the restriction on working memory capacity is not specifically about memory storage per se, but more about the capacity for sustained, regulated attention in the presence of interference.

\begin{figure*}[t]
    \begin{center}
         \begin{subfigure}{0.99\linewidth}
\includegraphics[width=0.95\linewidth]{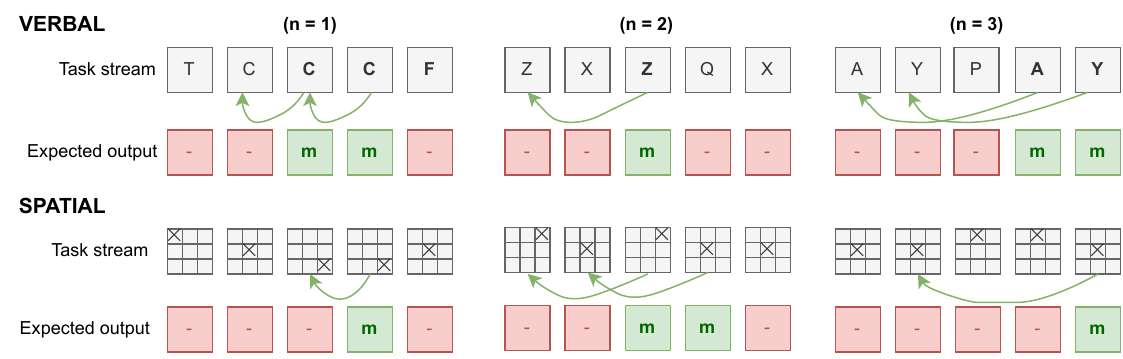}
    \end{subfigure}
    \end{center}
        \caption{
           Illustrations of verbal (top row) and spatial (bottom row) \textit{n}-back tasks with $n = \{1, 2, 3\}$. Participants are instructed to give a response (``\texttt{m}") when the current stimulus (e.g., a letter or a spatial location) is the same as the stimulus \textit{n} trial(s) ago, and not respond (``\texttt{-}") on nonmatch trials.
        }
        \label{fig:n_back}
\end{figure*}

\begin{figure}[ht!]
    \centering
    \includegraphics[width=0.25\textwidth]{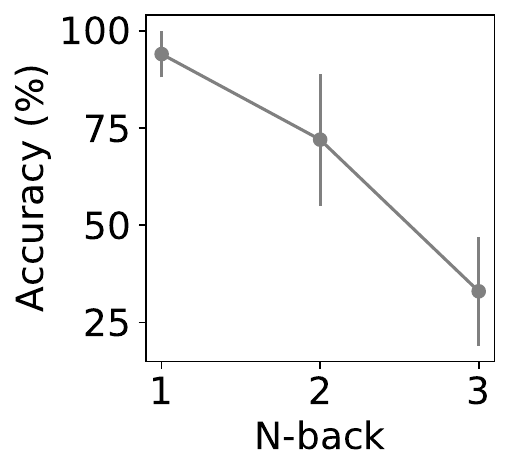}
    \caption{Typical human performance in \textit{n}-back tasks for $n=\{1, 2, 3\}$. We plot the mean $\pm1$ standard deviation of the data collected in \citet{jaeggi2010concurrent}.
    }
    \label{fig:human_nback}
\end{figure}

Supporting evidence of the executive attention hypothesis includes results from the \textit{n}-back task, which is arguably the gold-standard measure of working memory capacity in cognitive science~\cite{kaneRolePrefrontalCortex2002}. The \textit{n}-back task, initially developed by \citet{kirchner1958age}, requires participants to monitor a continuous stream of stimuli and to decide for each stimulus whether it matches the one \textit{n} step(s) back in the stream (see Figure~\ref{fig:n_back}). Participants in this task must, therefore, continuously update their mental representation of the target items while also dropping now irrelevant items from consideration. So, some executive attention processes are required in addition to storage. In this task, the level of \textit{n} at which a person's performance drop significantly can be taken as a measure of their working memory capacity. Typical human performance drops significantly when $n = 3$~\cite{klatzky2008n,amon2018auditory,jaeggi2010concurrent}, which can be defined as the working memory capacity limit of an average human. To illustrate this, we plot the data from one experiment presented in \citet{jaeggi2010concurrent} (see Figure~\ref{fig:human_nback}).

In humans, working memory capacity has proved to be closely related to fluid intelligence~\cite{cochraneFluidIntelligenceRelated2019,salthouseWhyWorkingMemory2008}, which refers to the ability to reason and to solve new problems independently of previously acquired knowledge. Training on working memory capacity using the \textit{n}-back task has been shown to be effective in improving fluid intelligence~\cite{auImprovingFluidIntelligence2015,jaeggiImprovingFluidIntelligence2008}, highlighting the special role of working memory capacity in human intelligence~\cite{halfordSeparatingCognitiveCapacity2007}. However, in artificial intelligence, there has not been a consensus as to which metrics should be accepted as an intelligence index when evaluating and comparing cognitive abilities of LLMs~\cite{doi:10.1126/science.adj5957}. In the current study, we define the working memory of LLMs as an emergent ability to selectively maintain and manipulate information for ongoing cognitive processes, and hypothesize that LLMs also have limited working memory capacity. Taking a step further, just as how critical working memory capacity is to human intelligence, it might also be used as an index of the intelligence emgerged from LLMs.

To investigate these hypotheses, we use ChatGPT (\texttt{gpt-3.5-turbo}) as a representative of LLMs and design two categories of \textit{n}-back tasks to evaluate its working memory capacity, which reveals strikingly consistent patterns of a capacity limit across multiple experimental conditions. We then compare the working memory capacity of different LLMs, and confirm that our proposed metric could be a strong correlate of the general capability of LLMs.

%% file: 3-background.tex
\section{Related Work}

Working memory has long been a subject of study in human cognition~\cite{cowanGeorgeMillerMagical2015}. Unlike long-term memory, which is stored in long-term synaptic weights, working memory is believed to be maintained by the activation of neurons in distributed brain networks~\cite{mejiasMechanismsDistributedWorking2022a}. However, the concept of working memory is largely unexplored in language models, with a few latest studies suggesting that imitating human-like working memory strategy can contribute to better performance of these models~\cite{guo2020working,li2022large}. Related to this, a better-known concept in LLMs is called in-context learning~\cite{brown2020language}, which demonstrates the ability of LLMs to retrieve long-term (pre-trained) knowledge and integrate the correct knowledge with the context, bearing a resemblance to how human working memory works. Previous studies have presented various approaches to leverage the in-context learning ability of language models, including selecting labeled examples for demonstrations~\cite{rubin2021learning,lu2021fantastically,liu2021makes}, meta-training with an explicit in-context learning objective ~\cite{chen2021meta,min2021metaicl}, and exploring the variant of in-context learning that involves learning to follow instructions~\cite{wei2022chain,wei2021finetuned,efrat2020turking,mishra2021reframing,mishra2021cross}.

However, to the best of our knowledge, this study is the first that provides an empirical analysis of the working memory capacity of LLMs from a cognitive science perspective.

%% file: 4-methodology.tex
\section{Methods}
We devised two categories of \textit{n}-back tasks involving verbal and spatial working memory~\cite{szmalec2011control} respectively and prompted ChatGPT (using the OpenAI API, model = ``\texttt{gpt-3.5-turbo}", temperature = $1$, other parameters are set to default values) to complete the tasks in a trial-by-trial manner. For both categories, we have a base version task and several variants derived from the base version further to test the model's performance under different conditions. To compare the performance of ChatGPT with other LLMs, we also used API of the following LLMs to perform the base version of the verbal task: \{\texttt{Bloomz-7B, Bloomz-7B1-mt, ChatGLM-6B\_v1.0, ChatGLM-6B\_v1.1, GPT-4, Vicuna-7B, Vicuna-13B}\}. All code for our experiments can be accessed in this repository: \url{https://github.com/Daniel-Gong/ChatGPT-WM}.

\begin{figure*}[t]
    \begin{center}
         \begin{subfigure}{0.99\linewidth}
\includegraphics[trim={0 0 0.2cm 0},clip,width=0.99\linewidth]{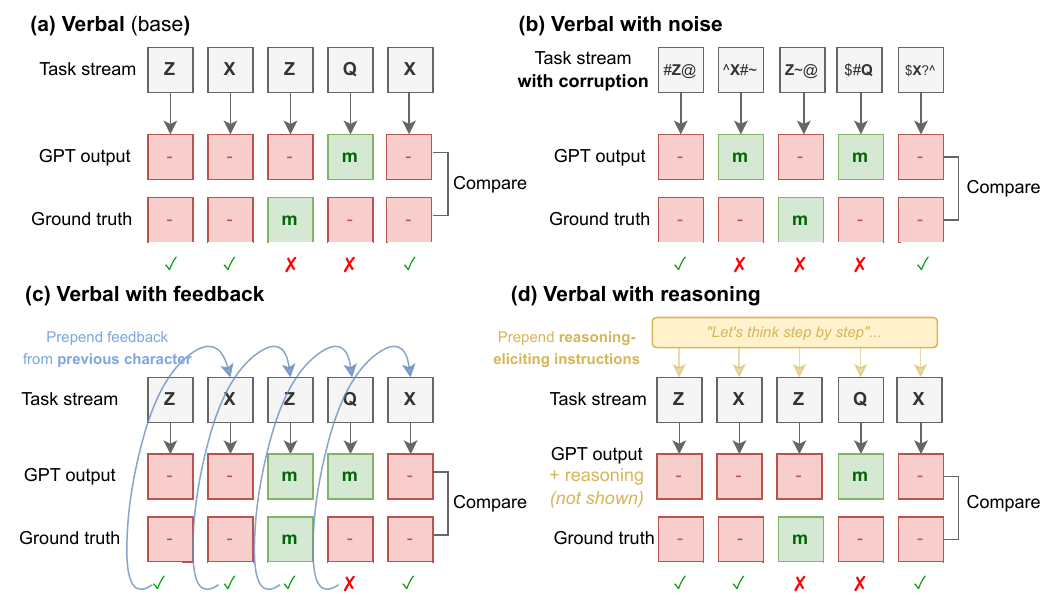}
    \end{subfigure}
    \end{center}
        \caption{
        Illustrations of the variants of verbal \textit{n}-back tasks (also applicable to spatial tasks). We use $n=2$ in the figure. (a): base version identical to the case presented in Figure \ref{fig:n_back} (top row); (b): stimulus on each trial now contains 3-6 random noise characters (chosen from ``\texttt{\#\$\%\&@\^{}\~{}}") in addition to a single alphabetical letter that the LLM should compare across trials. The LLM is instructed to ignore these noise characters, and the alphabetical letter may appear in any position in the noise-corrupted stimulus; (c): alongside the input for every trial, the LLM is also provided with feedback on whether it has performed the previous trial correctly; (d): the LLM is prompted with a reasoning-eliciting instruction to output the final answer (``\texttt{m}" or ``\texttt{-}") \textit{and} the rationale. Refer to Table \ref{tab:verbal_prompts} for the detailed prompts.
        }
    \label{fig:verbal_schematics}
\end{figure*}

\input{verb_prompt}

\paragraph{Verbal \textit{n}-back experiments.}
In the base version of the verbal \textit{n}-back task (see Figure \ref{fig:verbal_schematics}a), for $n = \{1, 2, 3\}$, respectively, we generated 50 blocks of letter sequences using an alphabet commonly found in the literature (``\texttt{bcdfghjklnpqrstvwxyz}"). Each block contained a sequence of 24 letters, which are presented one at a time as user input to the API. We included 8 match trials and 16 nonmatch trials in each block. The LLM was instructed to respond with ``\texttt{m}" on match trials and ``\texttt{-}" on nonmatch trials. Apart from the above base version, we further explored the behavioral performance of ChatGPT on the following three variants of the task:

\begin{itemize}
    \item We added $3$ to $6$ noise symbols to the input on every trial to examine the LLM's behavior when it is impossible to get the correct answer by simply doing a string match between stimulus inputs (see Figure \ref{fig:verbal_schematics}b).
    \item In human behavioral studies, a common strategy to improve participants' performance is to provide feedback after each trial~\cite{shalchyNBackRelatedERPs2020}. Here in the variant, after the LLM gave a response for the current trial, we provided feedback on whether its response was correct or wrong alongside the stimulus input of the following trial (see Figure \ref{fig:verbal_schematics}c).
    \item Chain-of-thought (CoT) prompting has proved helpful in eliciting reasoning in LLMs~\cite{wei2022chain}. In this variant, we instructed the LLM to think step by step when giving a response (see Figure \ref{fig:verbal_schematics}b).
\end{itemize}

\begin{figure*}[ht!]
    \begin{center}
         \begin{subfigure}{0.99\linewidth}
\includegraphics[trim={0 0 0.2cm 0},clip,width=0.99\linewidth]{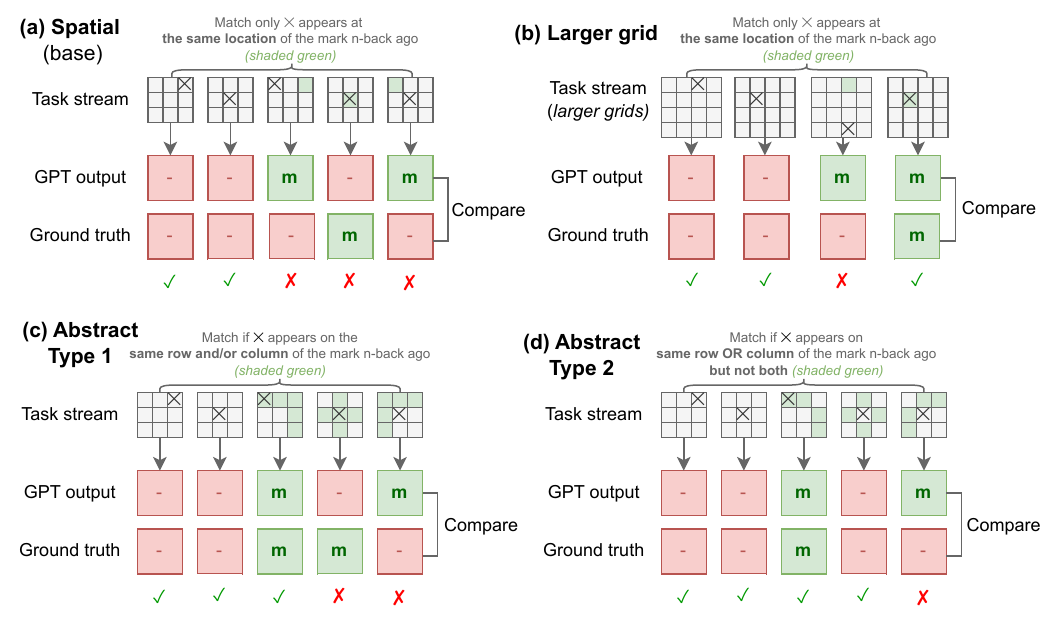}
    \end{subfigure}
    \end{center}
        \caption{
        Illustrations of the variants of spatial \textit{n}-back tasks ($n=2$ in the figure) besides those presented in Figure \ref{fig:verbal_schematics}. (a): base version identical to the case presented in Figure \ref{fig:n_back} (bottom row); (b): spatial tasks with larger grid sizes ($4 \times 4$ shown for illustration; we considered $4 \times 4$, $5 \times 5$, and $7 \times 7$); (c) and (d): two types of spatial reasoning tasks that additionally require \textit{abstract reasoning}. In (c), a match is expected whenever the letter X occurs in the same row and/or column as the location \textit{n} trials ago (including identical locations); in (d), a match is expected when the letter X appears in the same row or column (but not both) as the location \textit{n} trials ago (excluding identical locations). Refer to Table \ref{tab:spatial_prompts} for detailed prompts.
        }
        \label{fig:spatial_schematics}
\end{figure*}

\paragraph{Spatial \textit{n}-back experiments.}
Although in its very nature, LLMs are text-based, at least one study has demonstrated that they have spatial reasoning abilities~\cite{bubeck2023sparks}. To build on this promising trail and further examine the spatial working memory of ChatGPT, in the base version of the spatial \textit{n}-back task (Figure \ref{fig:spatial_schematics}a), we constructed a  $3 \times 3$ grid using ASCII characters. For $n = \{1, 2, 3\}$, respectively, we generated 50 blocks of grid sequences, each grid featuring a letter \textbf{X} in one of the nine positions. Note that the letter \textbf{X} was arbitrarily chosen to represent an occupied spatial location textually and could be substituted by any other letter or symbol. Each block contains 24 grids, including 8 match trials and 16 nonmatch trials. Like in the verbal \textit{n}-back tasks, the LLM was instructed to respond with ``\texttt{m}" on match trials and ``\texttt{-}" on nonmatch trials. We further explored the spatial working memory capacity of ChatGPT with the following modifications of the task:

\begin{itemize}
    \item Similar to the variants of verbal \textit{n}-back tasks, we also had ``spatial-with-noise", ``spatial-with-feedback", and ``spatial-with-CoT-reasoning" versions of the task. The with-feedback and with-CoT-reasoning variants were basically the same as those for the corresponding verbal tasks. For the spatial-with-noise version, we added a noise character (chosen from ``\texttt{\#\$\%\&@\^{}\~{}}") to $1$ to $3$ unoccupied locations in the 3 $\times$ 3 grid on every trial, so that we could examine the LLM's spatial working memory when it is not able to get the correct answer by simply doing string match.
    \item To test if the LLM can reason in a more sophisticated way, we further introduced two variants that specifically require abstract spatial reasoning. For the first variant (see Figure \ref{fig:spatial_schematics}c), a match is defined as when the location of the letter \textbf{X} is in the same row \textbf{and/or} column (i.e., including identical locations) as the \textbf{X} \textit{n} trials ago. For a second variant (see Figure \ref{fig:spatial_schematics}d), a match is defined as when the letter \textbf{X} appears in the same row \textbf{or} column, but not both (i.e., excluding identical locations). This constraint would further force the LLM to use abstract reasoning and instruction-following abilities to perform this task. Given the increased difficulty of these two variants, we expect the LLM would have a worse performance on these two variants compared to other variants.
    \item We also explored whether the size of the grid ($3 \times 3$, $4 \times 4$, $5 \times 5$ or $7 \times 7$) would influence the LLM's performance (see Figure \ref{fig:spatial_schematics}b). To the best of our knowledge, there have not been human studies exploring how the number of all possible spatial locations would impact behavioral performance in spatial \textit{n}-back tasks. In light of this, we did not have specific assumptions for how the LLM would perform differently under these scenarios.
\end{itemize}

\input{spatial_prompt}

%% file: verb_prompt.tex
\begin{table*}[ht]
\centering
   \renewcommand{\arraystretch}{1.5}
  \begin{tabular}{p{0.21\textwidth} p{0.72\textwidth}}
    \hline
    \textbf{Task type} & \textbf{Prompt} \\
    \hline
    Verbal \newline Verbal with Noise \newline Verbal with Feedback \newline (Figure \ref{fig:verbal_schematics}a-c) & You are asked to perform a \{1,2,3\}-back task. You will see a sequence of letters. The sequence will be presented one letter at a time, [\textbf{For the with-noise variant only:} accompanied with random noise symbols chosen from ``\texttt{\#\$\%\&@\^{}\~{}}". Please ignore the noise symbols and focus on the letter only]. Your task is to respond with ``\texttt{m}" whenever the current letter is the same as the previous \{one, two, three\} letter(s) ago, and ``\texttt{-}" otherwise. [\textbf{For the with-feedback variant only:} Feedback on whether your last response was correct or wrong will also be presented. Please take advantage of feedback information to improve your performance.] Only ``\texttt{m}" and ``\texttt{-}" are allowed responses. The sequence will be presented one letter at a time. Now begins the task. \\
    \hline
    Verbal with Reasoning (Figure \ref{fig:verbal_schematics}d) & You are asked to perform a \{1,2,3\}-back task. You will see a sequence of letters. The sequence will be presented one letter at a time. \newline
          Your task is to respond with ``\texttt{m}" whenever the current letter is the same as the letter \{one, two, three\} letter(s) ago, and ``\texttt{-}" otherwise.
          Please think step by step and provide your thinking steps after responding with ``\texttt{m}" or ``\texttt{-}". \newline
          Here are examples of how to format your response:\newline
          (1)``\texttt{-:this is the first trial, so my response is -}".\newline
          (2)``\texttt{m:the letter \{one, two, three\} trial(s) ago was a, the current letter is a, so my response is m}".\newline
          (3)``\texttt{-:the letter \{one, two, three\} letter(s) ago was a, the current letter is b, so my response is -}". \newline
          Now begins the task. \\
    \hline
  \end{tabular}
  \caption{Prompts used in different verbal task variants. Texts in curly brackets are selected as appropriate depending on the value of \textit{n}. Texts in square brackets are inserted as appropriate depending on task variants.}
  \label{tab:verbal_prompts}
\end{table*}

%% file: spatial_prompt.tex
\begin{table*}[ht]
\centering
\renewcommand{\arraystretch}{1.6}
  \begin{tabular}{p{0.2\textwidth} p{0.73\textwidth}}
    \hline
    \textbf{Task type} & \textbf{Prompt} \\
    \hline
    Spatial 3*3 Grid \newline (Figure \ref{fig:spatial_schematics}a) & You are asked to perform a \{1,2,3\}-back task. You will see a sequence of 3*3 grids. Each grid has a letter X in one of the nine positions. For example, a grid with X at top left corner would be \verb+``` |X|_|_| |_|_|_| |_|_|_| ```+. Your task is to respond with ``\texttt{m}" whenever the X is in the same position as \{the previous trial, two trials ago, three trials ago\}, and respond with ``\texttt{-}" otherwise. Only ``\texttt{m}" and ``\texttt{-}" are allowed responses. The sequence will be presented one grid at a time. Now begins the task. \\ \hline
    Spatial with Abstract Reasoning \newline (Figure \ref{fig:spatial_schematics}c-d) & You are asked to perform a \{1,2,3\}-back task. You will see a sequence of 3*3 grids. Each grid has a letter X in one of the nine positions. \newline For example, a grid with X at top left corner would be \verb+``` |X|_|_| |_|_|_| |_|_|_| ```+. Your task is to respond with ``\texttt{m}" whenever the X in the current grid is in the same row or column as the X \{in the previous trial, two trials ago, three trials ago\}, and ``\texttt{-}" otherwise. For example, the X in \verb+``` |X|_|_| |_|_|_| |_|_|_| ```+ is in the same row as the X in \verb+``` |_|X|_| |_|_|_| |_|_|_| ```+ and \verb+``` |_|_|X| |_|_|_| |_|_|_| ```+, and in the same column as the X in \verb+``` |_|_|_| |X|_|_| |_|_|_| ```+ and \verb+``` |_|_|_| |_|_|_| |X|_|_| ```+. [\textbf{For Type 1 only:} Note that \verb+``` |X|_|_| |_|_|_| |_|_|_| ```+ is also in the same row and column as \verb+``` |X|_|_| |_|_|_| |_|_|_| ```+ itself.] [\textbf{For Type 2 only:} Note that if the X \{in the previous trial, two trials ago, three trials ago\} was at the identical location to the X in the current grid, that does not count as a match: for example, \verb+``` |X|_|_| |_|_|_| |_|_|_| ```+ is not a match to \verb+``` |X|_|_| |_|_|_| |_|_|_| ```+ itself.] The sequence will be presented one grid at a time. Only ``\texttt{m}" and ``\texttt{-}" are allowed responses. Now begins the task.
 \\
    \hline
  \end{tabular}
  \caption{Prompts used in different spatial task variants. Texts in curly brackets are selected as appropriate depending on the value of \textit{n}. Texts in square brackets are inserted as appropriate depending on task variants. For prompts used in spatial-with-larger-grids, spatial-with-noise, spatial-with-feedback, and spatial-with-CoT-reasoning variants, refer to the code repository for details.}
  \label{tab:spatial_prompts}
\end{table*}

%% file: 5-experiments.tex
\section{Results}
To analyze the model's performance in our experiments, we used four widely accepted performance metrics reported in numerous human behavioral studies:  
\textbf{hit rate}, \textbf{false alarm rate}, \textbf{accuracy} and \textbf{detection sensitivity}.

In the current study, we did 50 blocks of tests for $n = \{1, 2, 3\}$ in each experiment, which allows us to calculate the standard error of the mean (\textit{SEM}) and draw error bars to visualize the reliability of our findings. Among the four metrics, the pattern of hit rates and false alarm rates can vary a lot depending on the specific task condition~\cite{chooiChangesErrorPatterns2020a}. Accuracy, in turn, will also be biased by very high/low hit rates and false alarm rates. In contrast, detection sensitivity($d'$) is a much more robust performance metric. A higher $d'$ indicates better performance, suggesting that the individual is more accurately distinguishing between targets and non-targets. Based on the overall difficulty of the current task, we set $d' = 1$ as the threshold to determine the working memory capacity of a model: if, at a certain level of $n$, the model's $d'$ drops to around $1$, we can define that its working memory capacity is limited to $n$. In light of this, our analysis below will mainly focus on $d'$ (see Appendix for the statistical tests we conducted).

\begin{figure*}[htb!]
    \begin{center}
         \begin{subfigure}{0.85\linewidth}
\includegraphics[trim={0 0 0.2cm 0},clip,width=0.99\linewidth]{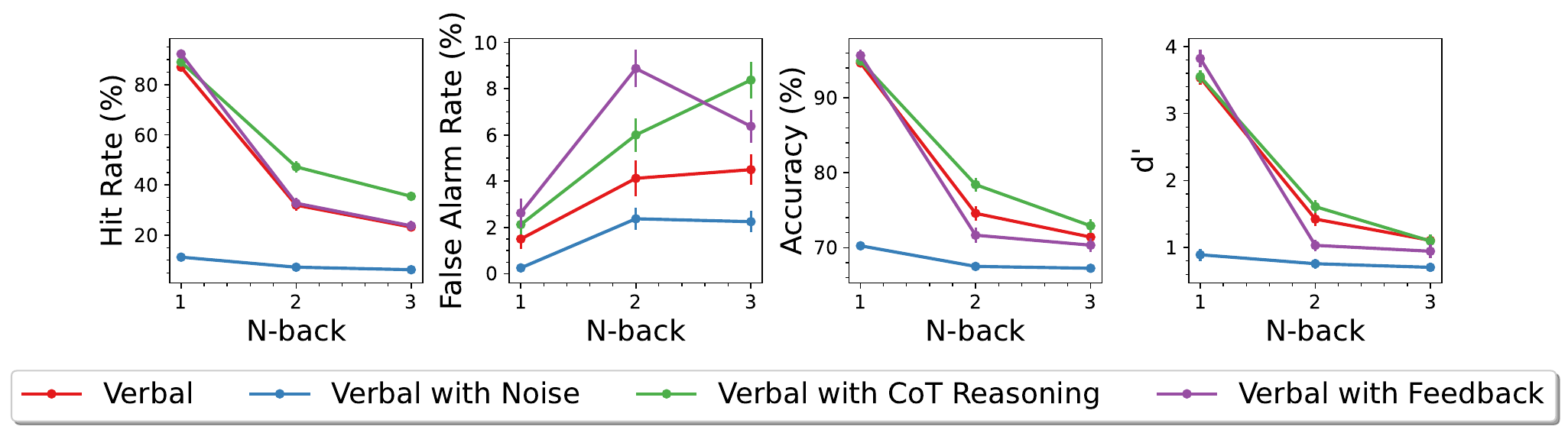}
    \end{subfigure}
    \end{center}
        \caption{
        Results of different variants of verbal \textit{n}-back experiments. Error bars represent $\pm1$ \textit{SEM}.
        }
        \label{fig:verbal_results}
\end{figure*}

\begin{figure*}[htb!]
    \begin{center}
         \begin{subfigure}{0.85\linewidth}
\includegraphics[trim={0 0 0.2cm 0},clip,width=0.99\linewidth]{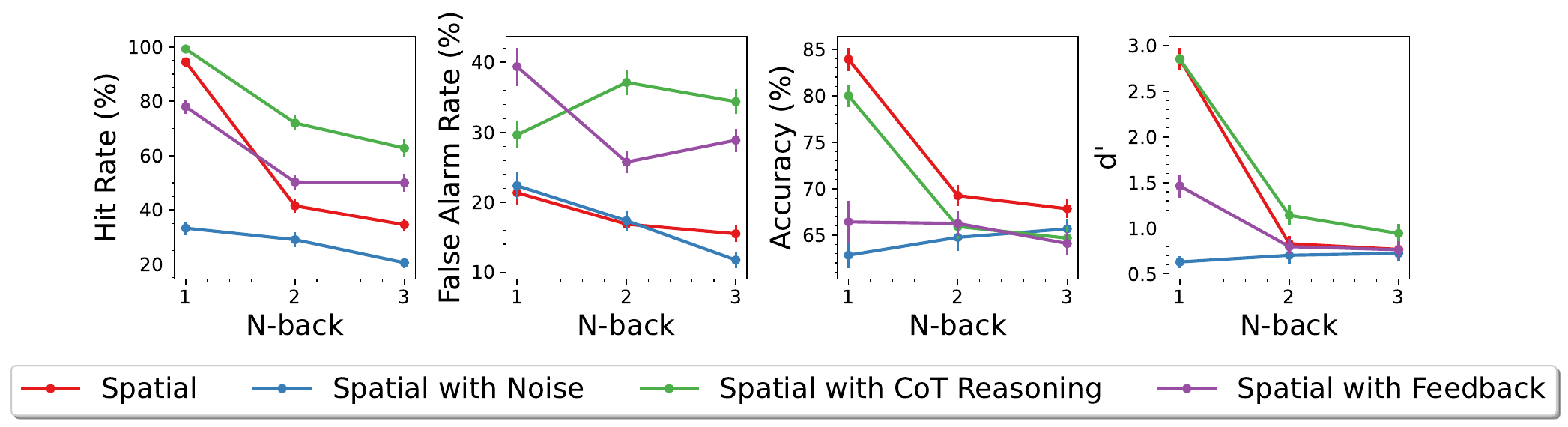}
    \end{subfigure}
    \end{center}
        \caption{
        Results of the variants of spatial \textit{n}-back tasks corresponding to those in verbal tasks. Error bars represent $\pm1$ \textit{SEM}.
        }
        \label{fig:spatial_results_1}
\end{figure*}

\begin{figure*}[htb!]
    \begin{center}
         \begin{subfigure}{0.98\linewidth}
\includegraphics[trim={0 0 0.2cm 0},clip,width=0.99\linewidth]{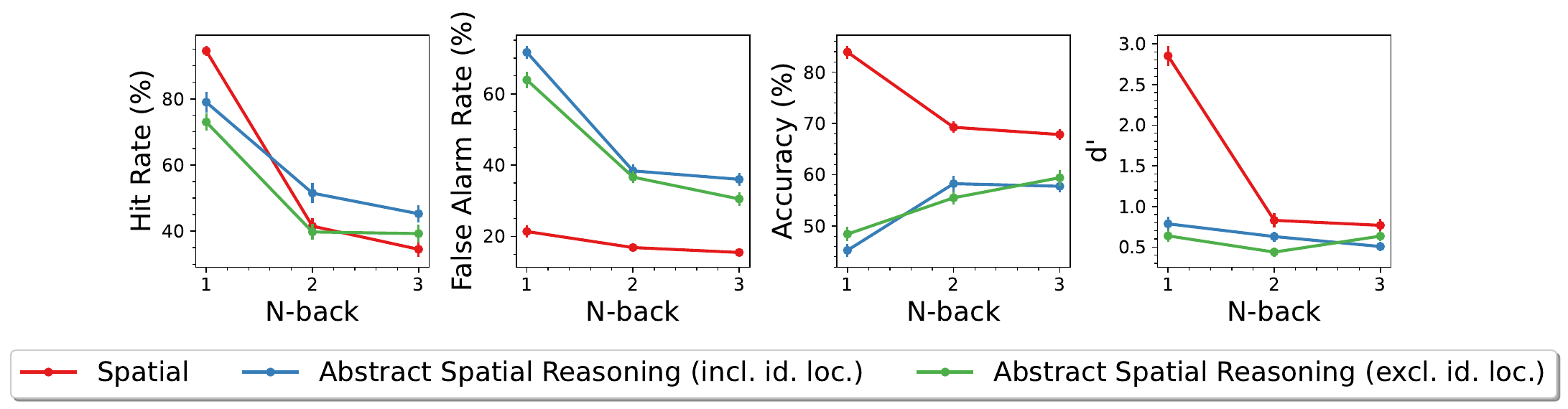}
    \end{subfigure}
    \end{center}
        \caption{
        Results of abstract reasoning variants of spatial \textit{n}-back tasks. Error bars represent $\pm1$ \textit{SEM}.
        }
        \label{fig:spatial_results_2}
\end{figure*}

\begin{figure*}[htb!]
    \begin{center}
         \begin{subfigure}{0.88\linewidth}
\includegraphics[trim={0 0 0.2cm 0},clip,width=0.99\linewidth]{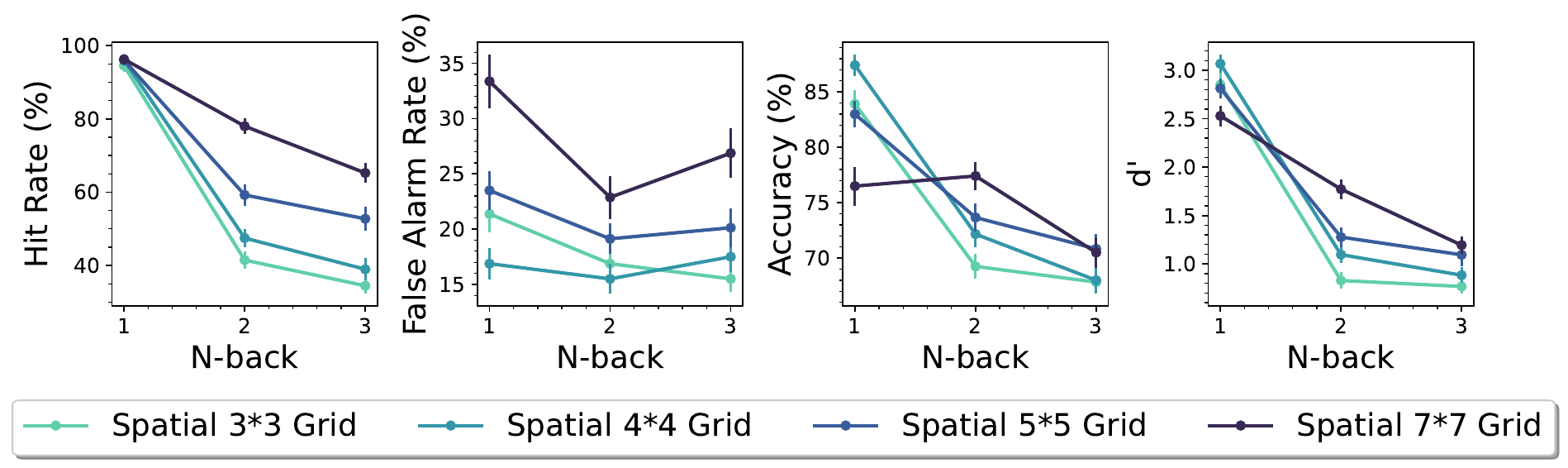}
    \end{subfigure}
    \end{center}
        \caption{
        Results of spatial task variants with different grid sizes. Error bars represent $\pm1$ \textit{SEM}.
        }
        \label{fig:spatial_results_3}
\end{figure*}

\begin{figure*}[htb!]
    \begin{center}
         \begin{subfigure}{0.85\linewidth}
\includegraphics[trim={0 0 0.2cm 0},clip,width=0.99\linewidth]{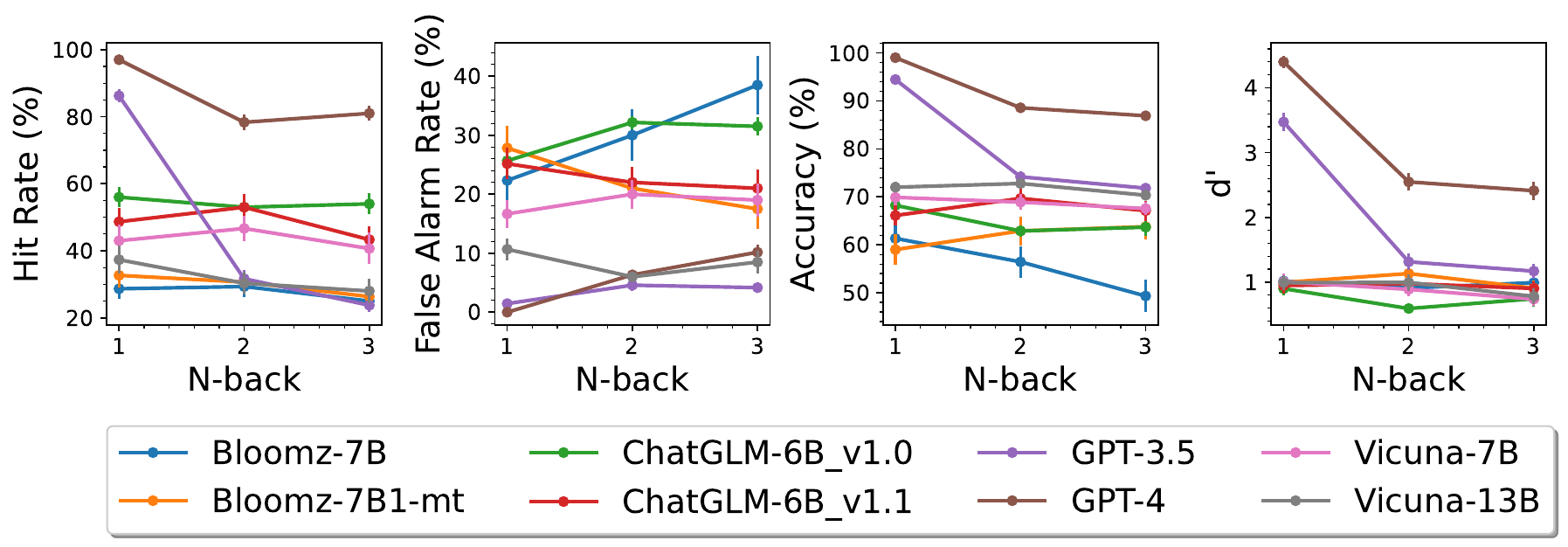}
    \end{subfigure}
    \end{center}
        \caption{
        Results of the verbal $n$-back task (base version) on different models. Error bars represent $\pm1$ \textit{SEM}.
        }
        \label{fig:verbal_all_models}
\end{figure*}

\paragraph{Verbal \textit{n}-back experiments.}
In the verbal task variants, we observed a performance pattern strikingly consistent with human participants, with the LLM's performance declining significantly when \textit{n} increased from 1 to 3 (Figure \ref{fig:verbal_results}). Furthermore, apart from the noise variant, all of the other three variants have a working memory capacity of around 3: their $d'$ drops to around $1$ when $n=3$. Adding noise significantly reduces the model's working memory capacity, which is analogous to distracting stimuli presented in human working memory experiments~\cite{doi:10.1073/pnas.1523471113}.

\paragraph{Spatial \textit{n}-back experiments.}
In the four versions of spatial tasks corresponding to the above verbal tasks, the same patterns of performance declines were basically replicated (Figure \ref{fig:spatial_results_1}). CoT reasoning significantly improved model performance, although overall, the model has a lower working memory capacity in the spatial variants compared to their verbal counterparts. We attribute this to the higher difficulty of spatial \textit{n}-back tasks compared to the verbal ones.

We further evaluated whether the LLM could conduct abstract spatial reasoning. As expected, the working memory capacity of the model when doing abstract reasoning was significantly lower than the base version (Figure \ref{fig:spatial_results_2}). Although the abstract reasoning variants haven't been done in human studies, we would expect to see similar decreases in working memory capacity in humans because of the highly cognitively demanding reasoning processes.

Our explorations on the effect of the grid size on model performance yielded interesting results, too. The LLM has a higher working memory capacity when the grid size is larger, as seen from the $d'$ results in Figure \ref{fig:spatial_results_3}. One possibility is that when the grid size is larger, there might be less interference between stimulus inputs across trials so that the LLM can better keep track of the information flow without being confused. Future studies should try to explain this phenomenon in more detail, and analogous tasks on human participants should be done to test the generalizability.

\paragraph{Model comparison.}
To investigate whether other LLMs exhibit similar performance patterns, we tested 7 other LLMs on the base version of the verbal \textit{n}-back task (Figure~\ref{fig:verbal_all_models}). Strikingly, GPT-4, which is arguably the most intelligent LLM today, also possesses a working memory capacity that far exceeds that of other LLMs. However, due to the high cost of calling the API of GPT-4, we did not test it with $n > 3$ to determine its exact working memory capacity. In contrast, other open-source LLMs (Bloomz-7B, Bloomz-7B1-mt, ChatGLM-6B\_v1.0, ChatGLM-6B\_v1.1, Vicuna-7B, Vicuna-13B), which are considered less capable than GPT-3.5 and GPT-4, have a very low working memory capacity and are nearly indistinguishable from each other.

%% file: 6-conclusion.tex
\section{Discussion}

We discover that ChatGPT has limited working memory capacity, and that its capacity limit is similar to that of humans. Although some prompting techniques may be used to improve the model's performance, the trend of performance declines and the capacity limit still bear a striking resemblance to humans. This consistent pattern thus might be reflecting a fundamental constraint that emerged from the architecture of the model, suggesting a possibility that the low-level mechanisms of working memory in ChatGPT might be similar to human working memory, at least in some aspects.

Our model comparison results further confirm that the performance of LLMs on \textit{n}-back tasks can be a reliable metric for assessing their working memory capacity, which in turn might reflect the general intelligence of reasoning and problem-solving emerged from these models. Future studies should test LLMs on other working memory span tasks used in cognitive science~\cite{conwayWorkingMemorySpan2005a,danemanIndividualDifferencesWorking1980} to address the generalizability of \textit{n}-back tasks as measurement tools. Finally, exploring how the transformer architecture (especially self-attention mechanism) plays a role in the capacity limit would have the potential to improve the working memory capacity of these models, which in turn could boost the overall intelligence level of future LLMs.

%% file: 7-supplement.tex
\section*{Appendix}
\subsection{Statistical Tests}
\label{appendix:tests}
Due to the fact that the experimental data do not conform to the assumptions of parametric tests (normality and homogeneity of the variance), we used non-parametric Kruskal-Wallis H tests and reported \textit{H} value, \textit{p} value, and $\epsilon^2$ (effect size) to investigate if there is a significant difference in $d'$ across $n = \{1, 2, 3\}$. After that, we did non-parametric Wilcoxon signed rank tests and reported \textit{T} value (a smaller value of \textit{T} indicates a stronger deviation from the null hypothesis), \textit{p} value, and effect size (rank-biserial correlation, \textit{r}) to examine if the $d'$ when $n = \{1, 2, 3\}$ is significantly greater than $1$. Note that when comparing different LLMs' performance on the verbal \textit{n}-back task (base version), we did $30$ (instead of $50$) blocks of tests for $n = \{1, 2, 3\}$ in each experiment to shorten the time needed to complete the experiments.


\begin{table}[ht!]
\centering
\caption{Kruskal-Wallis H test statistics on \textbf{verbal} tasks.}
\begin{tabular}{lrrr}
\hline
 Task                      &        \textit{H} &           \textit{p} &   $\epsilon^2$ \\
\hline
 Verbal                    & 97.5376  & 6.60666e-22 &      0.649916  \\
 Verbal with Noise         &  3.91569 & 0.141162    &      0.0130319 \\
 Verbal with CoT Reasoning & 99.4143  & 2.58493e-22 &      0.662683  \\
 Verbal with Feedback      & 94.9077  & 2.46072e-21 &      0.632025  \\
\hline
\end{tabular}
\label{tab:h_verbal}
\end{table}

\begin{table}[ht!]
\centering
\caption{Wilcoxon signed rank test statistics on \textbf{verbal} tasks.}
\scalebox{0.85}{
\centering
\begin{tabular}{llrrr}
\hline
 Task                      & \textit{n}-back   &    \textit{T} &           \textit{p} &          \textit{r} \\
\hline
 Verbal                    & 1-back   & 1275 & 8.88178e-16 &  1         \\
 Verbal                    & 2-back   & 1046 & 1.9091e-05  &  0.640784  \\
 Verbal                    & 3-back   &  811 & 0.0475667   &  0.272157  \\
 Verbal with Noise         & 1-back   &  620 & 0.568367    & -0.027451  \\
 Verbal with Noise         & 2-back   &  299 & 0.999608    & -0.53098   \\
 Verbal with Noise         & 3-back   &  231 & 0.99998     & -0.637647  \\
 Verbal with CoT Reasoning & 1-back   & 1275 & 8.88178e-16 &  1         \\
 Verbal with CoT Reasoning & 2-back   & 1160 & 1.547e-08   &  0.819608  \\
 Verbal with CoT Reasoning & 3-back   &  719 & 0.218784    &  0.127843  \\
 Verbal with Feedback      & 1-back   & 1275 & 8.88178e-16 &  1         \\
 Verbal with Feedback      & 2-back   &  709 & 0.248194    &  0.112157  \\
 Verbal with Feedback      & 3-back   &  588 & 0.683913    & -0.0776471 \\
\hline
\end{tabular}
}
\label{tab:u_verbal}
\end{table}



\begin{table}[ht!]
\caption{Kruskal-Wallis H test statistics on \textbf{spatial} tasks corresponding to the verbal ones.}
\centering
\begin{tabular}{lrrr}
\hline
 Task                       &        \textit{H} &           \textit{p} &   $\epsilon^2$ \\
\hline
 Spatial                    & 84.9206  & 3.62842e-19 &     0.564086   \\
 Spatial with Noise         &  0.63338 & 0.728557    &    -0.00929674 \\
 Spatial with CoT Reasoning & 88.4591  & 6.18527e-20 &     0.588157   \\
 Spatial with Feedback      & 21.4206  & 2.23139e-05 &     0.132113   \\
\hline
\end{tabular}
\label{tab:h_spatial_verbal}
\end{table}

\begin{table}[ht!]
\centering
\caption{Wilcoxon signed rank test statistics on \textbf{spatial} tasks corresponding to the verbal ones.}
\scalebox{0.85}{
\centering
\begin{tabular}{llrrr}
\hline
 Task                       & \textit{n}-back   &   \textit{T} &           \textit{p} &          \textit{r} \\
\hline
 Spatial                    & 1-back   & 464 & 1.86265e-09 &  0.995699  \\
 Spatial                    & 2-back   & 158 & 0.937933    & -0.32043   \\
 Spatial                    & 3-back   & 139 & 0.973868    & -0.402151  \\
 Spatial with Noise         & 1-back   &  62 & 0.999906    & -0.733333  \\
 Spatial with Noise         & 2-back   &  54 & 0.99996     & -0.767742  \\
 Spatial with Noise         & 3-back   & 130 & 0.98364     & -0.44086   \\
 Spatial with CoT Reasoning & 1-back   & 465 & 9.31323e-10 &  1         \\
 Spatial with CoT Reasoning & 2-back   & 217 & 0.627173    & -0.0666667 \\
 Spatial with CoT Reasoning & 3-back   & 163 & 0.924057    & -0.298925  \\
 Spatial with Feedback      & 1-back   & 327 & 0.0261317   &  0.406452  \\
 Spatial with Feedback      & 2-back   & 159 & 0.935323    & -0.316129  \\
 Spatial with Feedback      & 3-back   & 178 & 0.868939    & -0.234409  \\
\hline
\end{tabular}
}
\label{tab:u_spatial_verbal}
\end{table}

\clearpage

\begin{table}[h]
\caption{Kruskal-Wallis H test statistics on the abstract reasoning variants of \textbf{spatial} tasks.}
\centering
\begin{tabular}{lrrr}
\hline
 Task                                                            &        \textit{H} &           \textit{p} &   $\epsilon^2$ \\
\hline
 Spatial                                                         & 84.9206  & 3.62842e-19 &      0.564086  \\
 Abstract Reasoning (incl. identical) &  4.06941 & 0.130719    &      0.0140776 \\
 Abstract Reasoning (excl. identical) &  7.19739 & 0.0273595   &      0.0353564 \\
\hline
\end{tabular}
\end{table}

\begin{table}[h]
\centering
\caption{Wilcoxon signed rank test statistics on the abstract reasoning variants of \textbf{spatial} tasks.}
\scalebox{0.85}{
\centering
\begin{tabular}{llrrr}
\hline
 Task                                              & \textit{n}-back   &   \textit{T} &           \textit{p} &         \textit{r} \\
\hline
 Spatial                                           & 1-back   & 464 & 1.86265e-09 &  0.995699 \\
 Spatial                                           & 2-back   & 158 & 0.937933    & -0.32043  \\
 Spatial                                           & 3-back   & 139 & 0.973868    & -0.402151 \\
 Spatial with Abstract Reasoning (incl. identical) & 1-back   & 172 & 0.893534    & -0.260215 \\
 Spatial with Abstract Reasoning (incl. identical) & 2-back   &  57 & 0.999945    & -0.754839 \\
 Spatial with Abstract Reasoning (incl. identical) & 3-back   &  37 & 0.999995    & -0.84086  \\
 Spatial with Abstract Reasoning (excl. identical) & 1-back   &  64 & 0.999884    & -0.724731 \\
 Spatial with Abstract Reasoning (excl. identical) & 2-back   &  33 & 0.999997    & -0.858065 \\
 Spatial with Abstract Reasoning (excl. identical) & 3-back   &  40 & 0.999993    & -0.827957 \\
\hline
\end{tabular}
}
\end{table}



\begin{table}[h]
\caption{Kruskal-Wallis H test statistics on the \textbf{spatial} task variants with different grid sizes.}
\centering
\begin{tabular}{lrrr}
\hline
 Task        &       \textit{H} &           \textit{p} &   $\epsilon^2$ \\
\hline
 Spatial 3*3 & 84.9206 & 3.62842e-19 &       0.564086 \\
 Spatial 4*4 & 93.9609 & 3.95043e-21 &       0.625585 \\
 Saptial 5*5 & 73.0433 & 1.37675e-16 &       0.483288 \\
 Spatial 7*7 & 53.6315 & 2.25977e-12 &       0.351235 \\
\hline
\end{tabular}
\end{table}

\begin{table}[h!]
\caption{Wilcoxon signed rank test statistics on the \textbf{spatial} task variants with different grid sizes.}
\centering
\begin{tabular}{llrrr}
\hline
 Task        & \textit{n}-back   &   \textit{T} &           \textit{p} &         \textit{r} \\
\hline
 Spatial 3*3 & 1-back   & 464 & 1.86265e-09 &  0.995699 \\
 Spatial 3*3 & 2-back   & 158 & 0.937933    & -0.32043  \\
 Spatial 3*3 & 3-back   & 139 & 0.973868    & -0.402151 \\
 Spatial 4*4 & 1-back   & 465 & 9.31323e-10 &  1        \\
 Spatial 4*4 & 2-back   & 260 & 0.291879    &  0.11828  \\
 Spatial 4*4 & 3-back   & 180 & 0.859957    & -0.225806 \\
 Saptial 5*5 & 1-back   & 465 & 9.31323e-10 &  1        \\
 Saptial 5*5 & 2-back   & 296 & 0.0990379   &  0.273118 \\
 Saptial 5*5 & 3-back   & 264 & 0.264553    &  0.135484 \\
 Spatial 7*7 & 1-back   & 462 & 4.65661e-09 &  0.987097 \\
 Spatial 7*7 & 2-back   & 437 & 1.38115e-06 &  0.87957  \\
 Spatial 7*7 & 3-back   & 276 & 0.190899    &  0.187097 \\
\hline
\end{tabular}
\end{table}

\clearpage

\begin{table}[htbp!]
\caption{Kruskal-Wallis H test statistics on the base version of the verbal tasks using \textbf{different models}.}
\centering
\begin{tabular}{lrrr}
\hline
 Model            &          \textit{H} &           \textit{p} &   $\epsilon^2$ \\
\hline
 Bloomz-7B       &  0.729608  & 0.694333    &   -0.0146022   \\
 Bloomz-7B1-mt   &  2.24773   & 0.325021    &    0.00284745  \\
 ChatGLM-6B\_v1.0 &  4.14489   & 0.125878    &    0.0246539   \\
 ChatGLM-6B\_v1.1 &  0.0159079 & 0.992078    &   -0.0228057   \\
 GPT-3.5         & 58.5421    & 1.93972e-13 &    0.64991     \\
 GPT-4           & 52.5136    & 3.95191e-12 &    0.580617    \\
 Vicuna-7B       &  2.68935   & 0.260624    &    0.00792362  \\
 Vicuna-13B      &  1.97187   & 0.37309     &   -0.000323345 \\
\hline
\end{tabular}
\end{table}

\begin{table}[htbp!]
\caption{Wilcoxon signed rank test statistics on the base version of the verbal tasks using \textbf{different models}.}
\centering
\begin{tabular}{llrrr}
\hline
 Model           & \textit{n}-back   &   \textit{T} &           \textit{p} &           \textit{r} \\
\hline
 Bloomz-7B       & 1-back   & 240 & 0.443597    &  0.0322581  \\
 Bloomz-7B       & 2-back   & 185 & 0.835765    & -0.204301   \\
 Bloomz-7B       & 3-back   & 218 & 0.619467    & -0.0623656  \\
 Bloomz-7B1-mt   & 1-back   & 209 & 0.686827    & -0.101075   \\
 Bloomz-7B1-mt   & 2-back   & 287 & 0.135503    &  0.234409   \\
 Bloomz-7B1-mt   & 3-back   & 181 & 0.855317    & -0.221505   \\
 ChatGLM-6B\_v1.0 & 1-back   & 168 & 0.908015    & -0.277419   \\
 ChatGLM-6B\_v1.0 & 2-back   &  21 & 1           & -0.909677   \\
 ChatGLM-6B\_v1.0 & 3-back   & 103 & 0.996903    & -0.556989   \\
 ChatGLM-6B\_v1.1 & 1-back   & 199 & 0.755077    & -0.144086   \\
 ChatGLM-6B\_v1.1 & 2-back   & 219 & 0.611716    & -0.0580645  \\
 ChatGLM-6B\_v1.1 & 3-back   & 176 & 0.877527    & -0.243011   \\
 GPT-3.5         & 1-back   & 465 & 9.31323e-10 &  1          \\
 GPT-3.5         & 2-back   & 351 & 0.00683162  &  0.509677   \\
 GPT-3.5         & 3-back   & 321 & 0.0349465   &  0.380645   \\
 GPT-4           & 1-back   & 465 & 9.31323e-10 &  1          \\
 GPT-4           & 2-back   & 465 & 9.31323e-10 &  1          \\
 GPT-4           & 3-back   & 465 & 9.31323e-10 &  1          \\
 Vicuna-7B       & 1-back   & 217 & 0.627173    & -0.0666667  \\
 Vicuna-7B       & 2-back   & 193 & 0.791935    & -0.169892   \\
 Vicuna-7B       & 3-back   & 119 & 0.991273    & -0.488172   \\
 Vicuna-13B      & 1-back   & 219 & 0.611716    & -0.0580645  \\
 Vicuna-13B      & 2-back   & 234 & 0.491917    &  0.00645161 \\
 Vicuna-13B      & 3-back   & 151 & 0.954025    & -0.350538   \\
\hline
\end{tabular}
\end{table}

\clearpage

\subsection{Performance Distributions}
\label{appendix:distributions}
To get a better sense of the task performance across blocks, below we plotted the distributions of $d'$ from all the tasks.


\begin{figure}[htb!]
\centering
\includegraphics[width=0.5\textwidth]{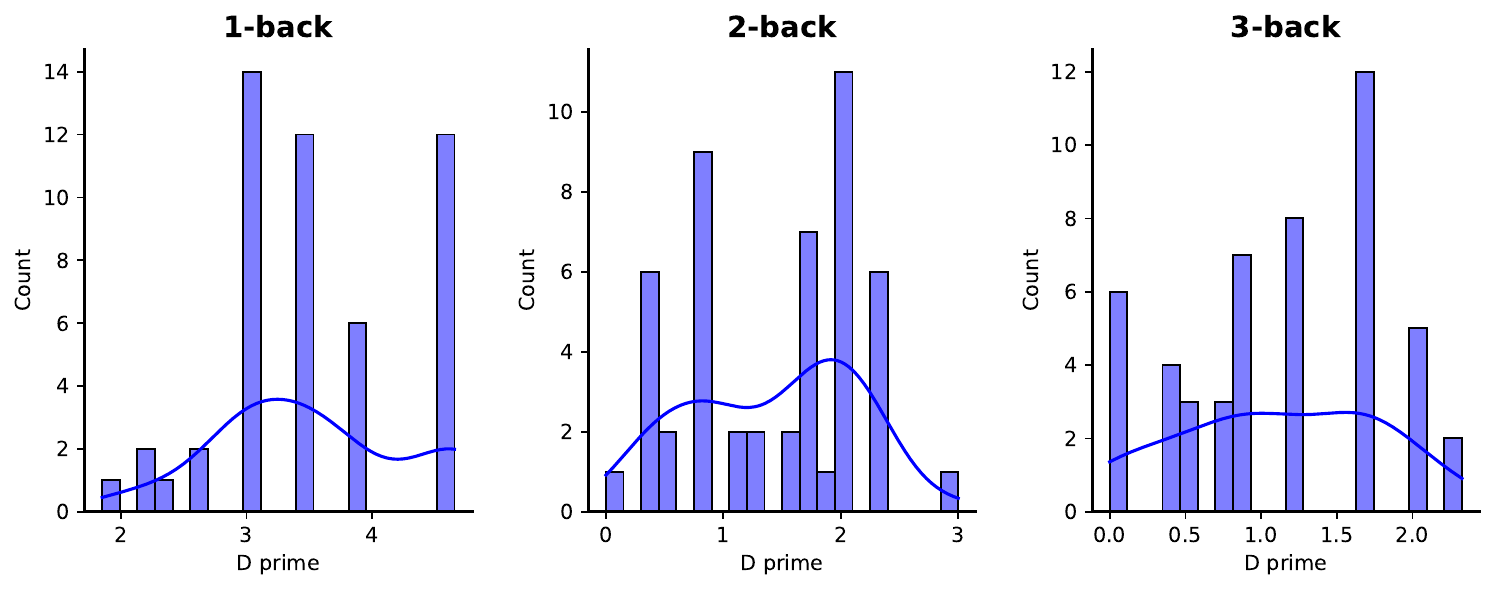} 
\caption{$d'$ distributions: verbal (base version).}
\end{figure}


\begin{figure}[htb!]
\centering
\includegraphics[width=0.5\textwidth]{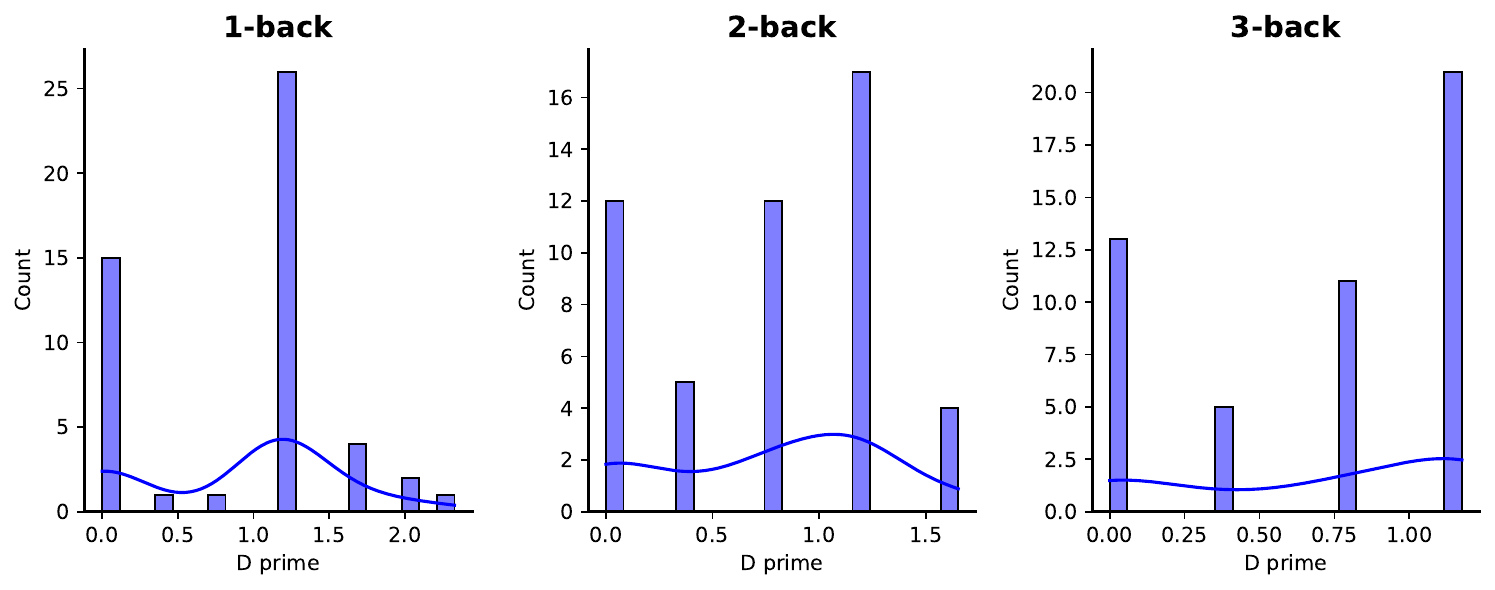} 
\caption{$d'$ distributions: verbal with noise.}
\end{figure}


\begin{figure}[htb!]
\centering
\includegraphics[width=0.5\textwidth]{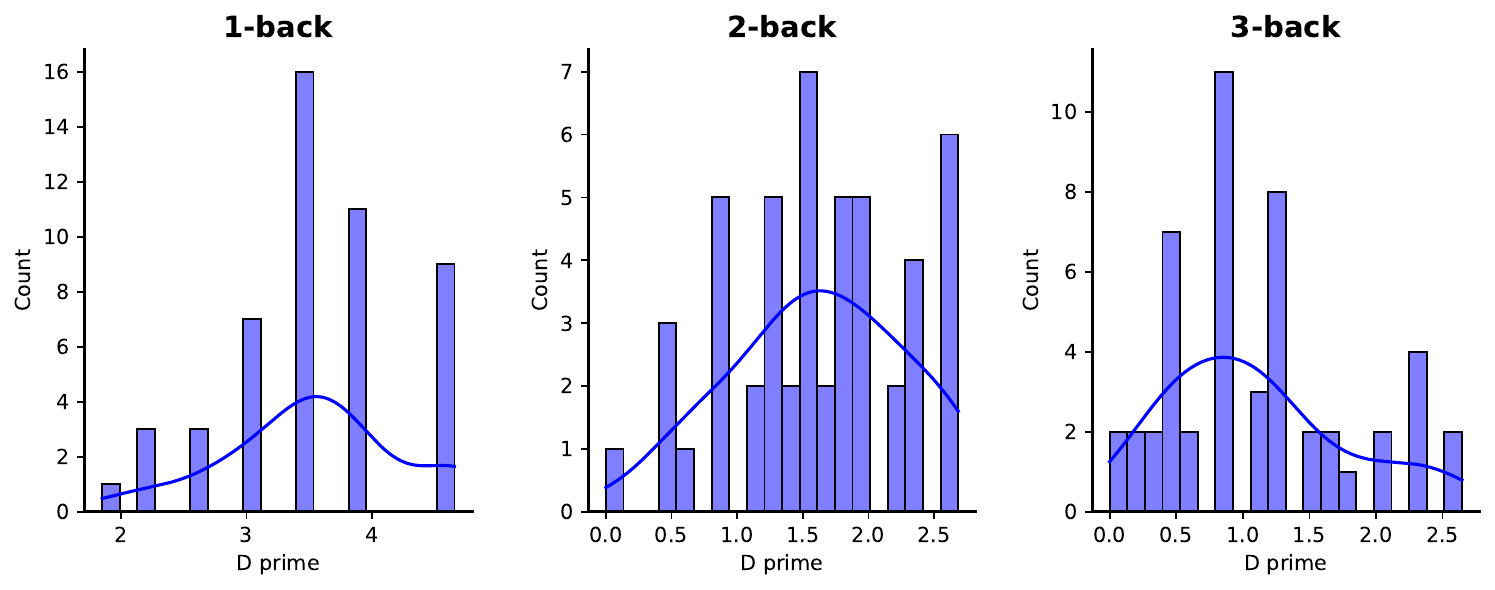} 
\caption{$d'$ distributions: verbal with CoT reasoning.}
\end{figure}


\begin{figure}[htb!]
\centering
\includegraphics[width=0.5\textwidth]{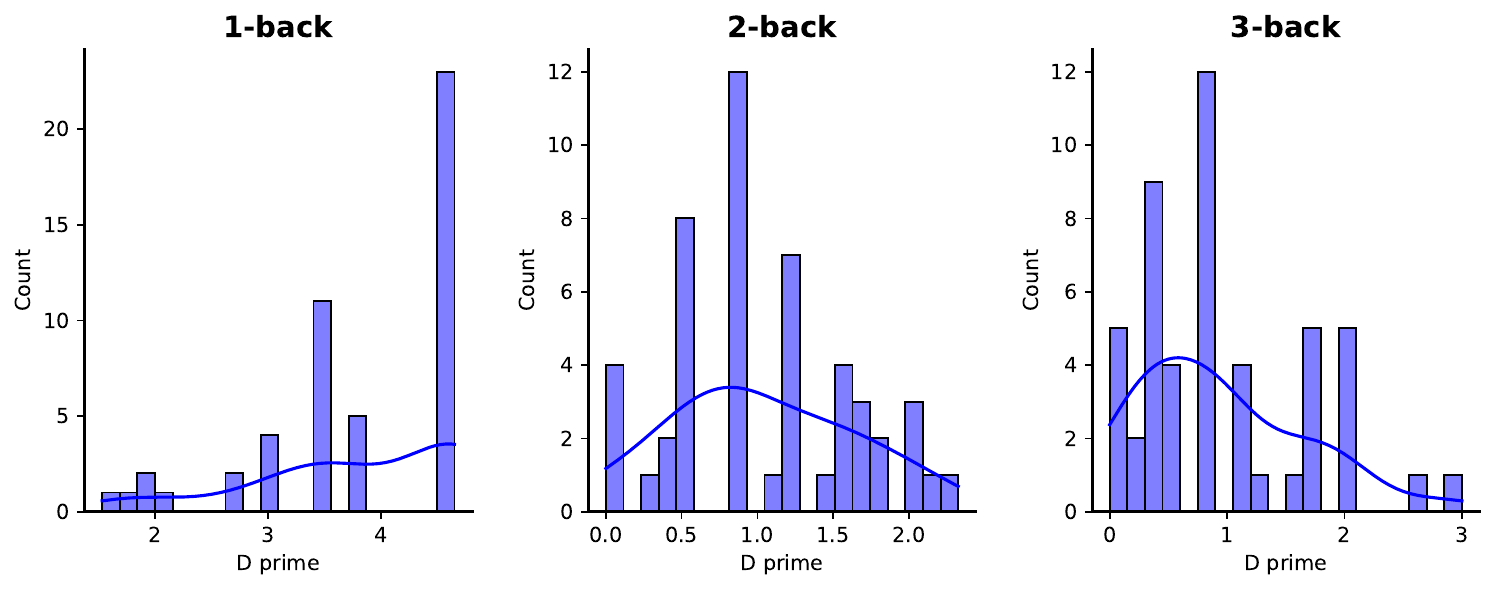} 
\caption{$d'$ distributions: verbal with feedback.}
\end{figure}


\begin{figure}[htb!]
\centering
\includegraphics[width=0.5\textwidth]{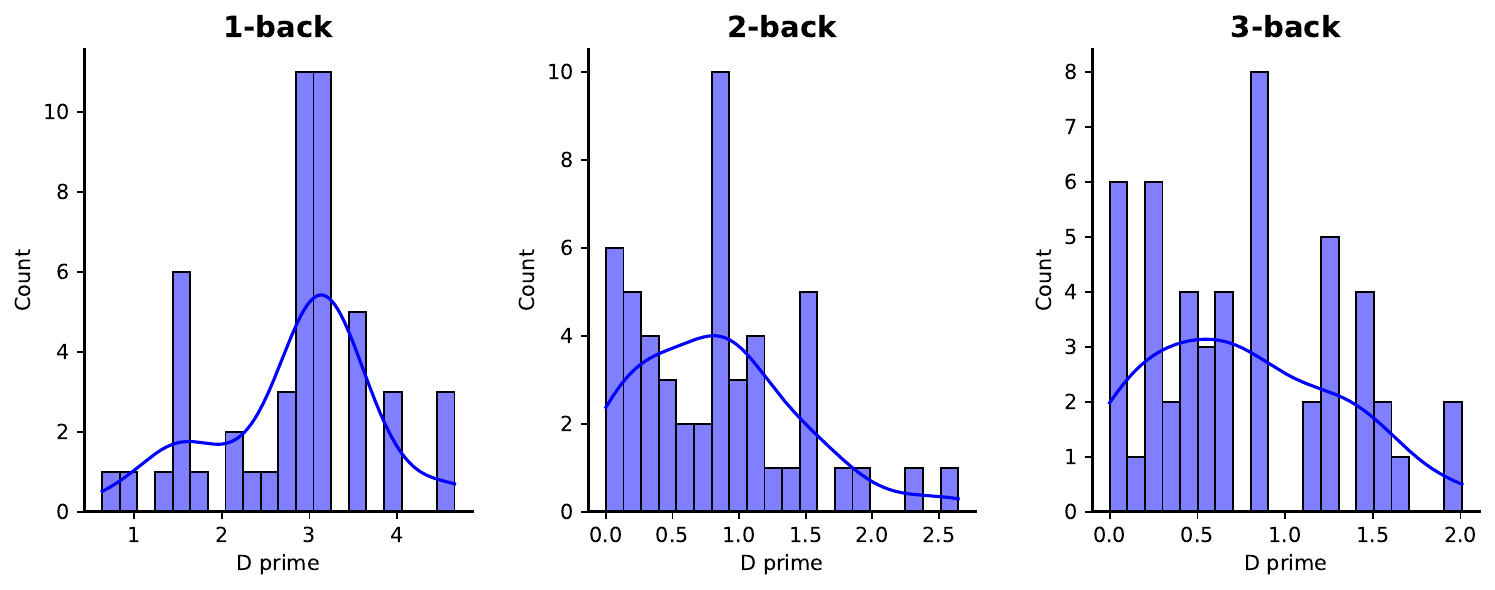} 
\caption{$d'$ distributions: spatial (base version).}
\end{figure}


\begin{figure}[htb!]
\centering
\includegraphics[width=0.5\textwidth]{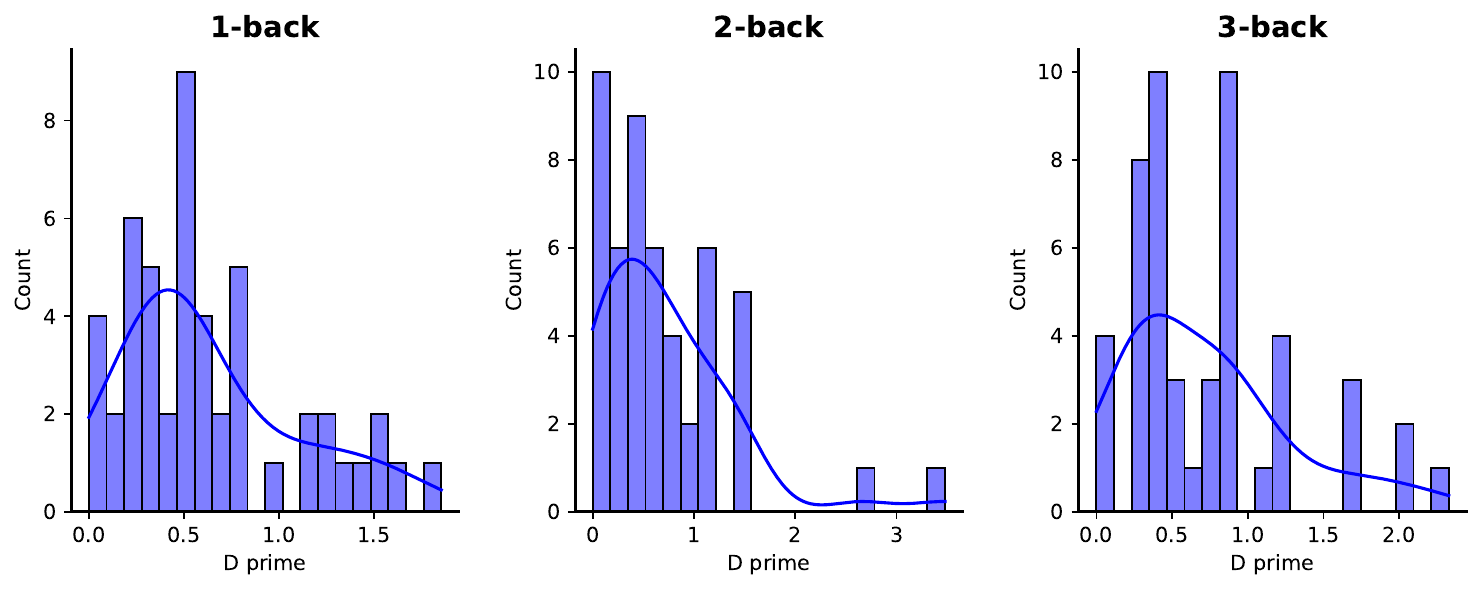} 
\caption{$d'$ distributions: spatial with noise.}
\end{figure}


\begin{figure}[htb!]
\centering
\includegraphics[width=0.5\textwidth]{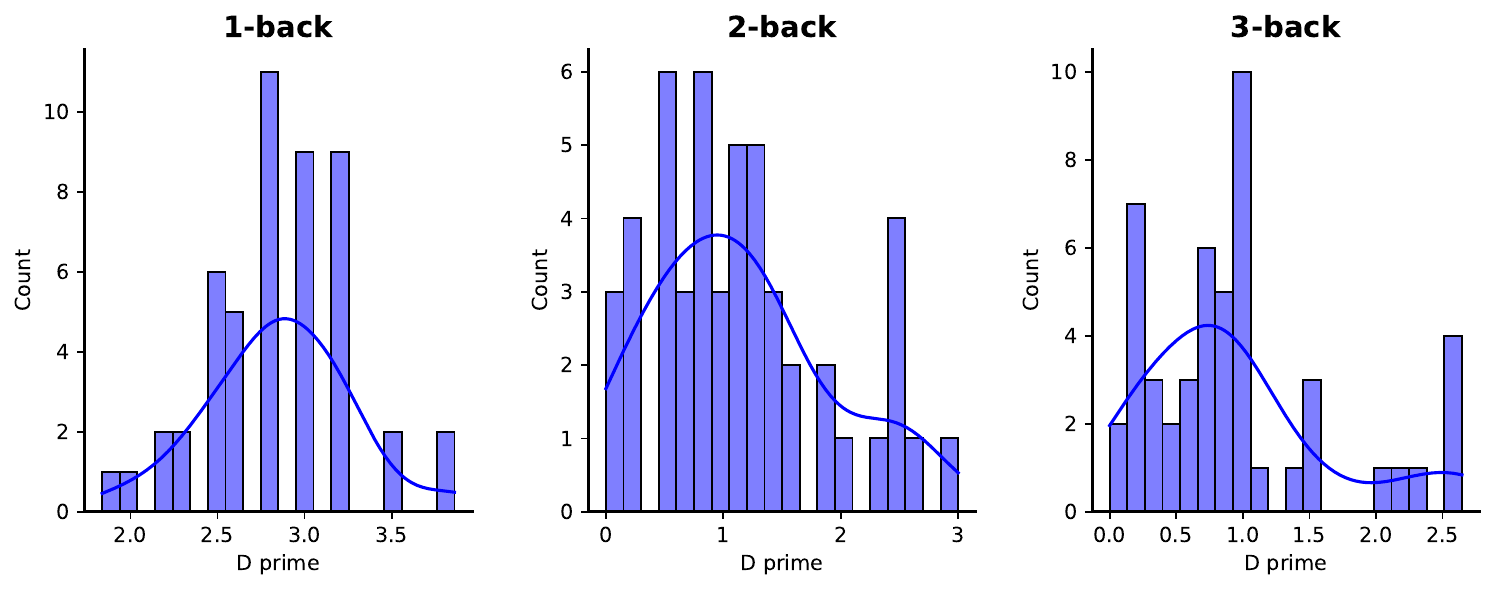} 
\caption{$d'$ distributions: spatial with CoT reasoning.}
\end{figure}


\begin{figure}[htb!]
\centering
\includegraphics[width=0.5\textwidth]{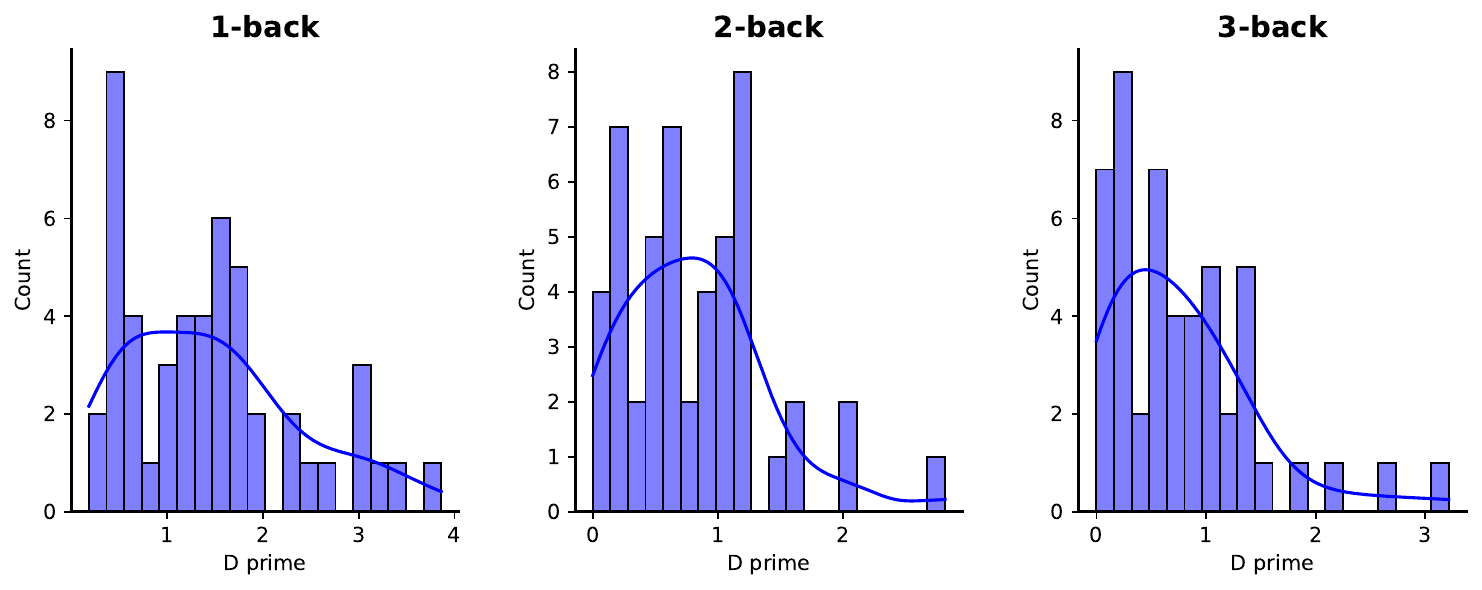} 
\caption{$d'$ distributions: spatial with feedback.}
\end{figure}


\begin{figure}[htb!]
\centering
\includegraphics[width=0.5\textwidth]{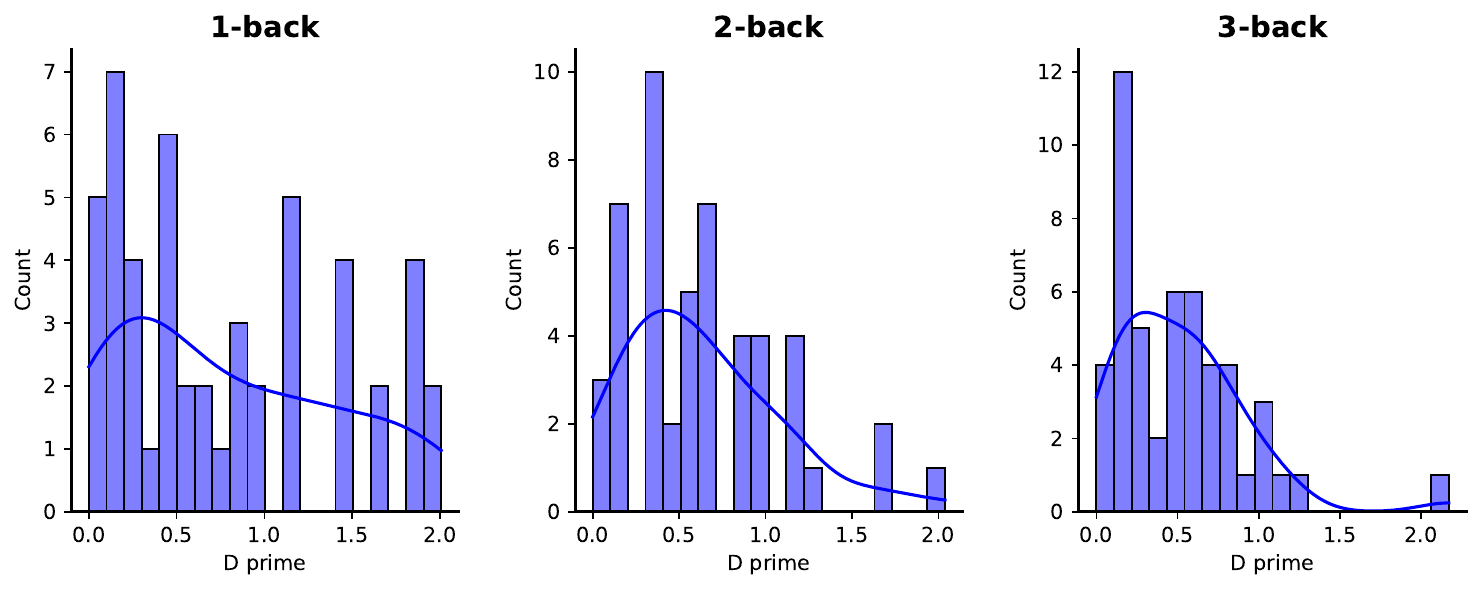} 
\caption{$d'$ distributions: spatial with abstract reasoning (including identical locations).}
\end{figure}


\begin{figure}[htb!]
\centering
\includegraphics[width=0.5\textwidth]{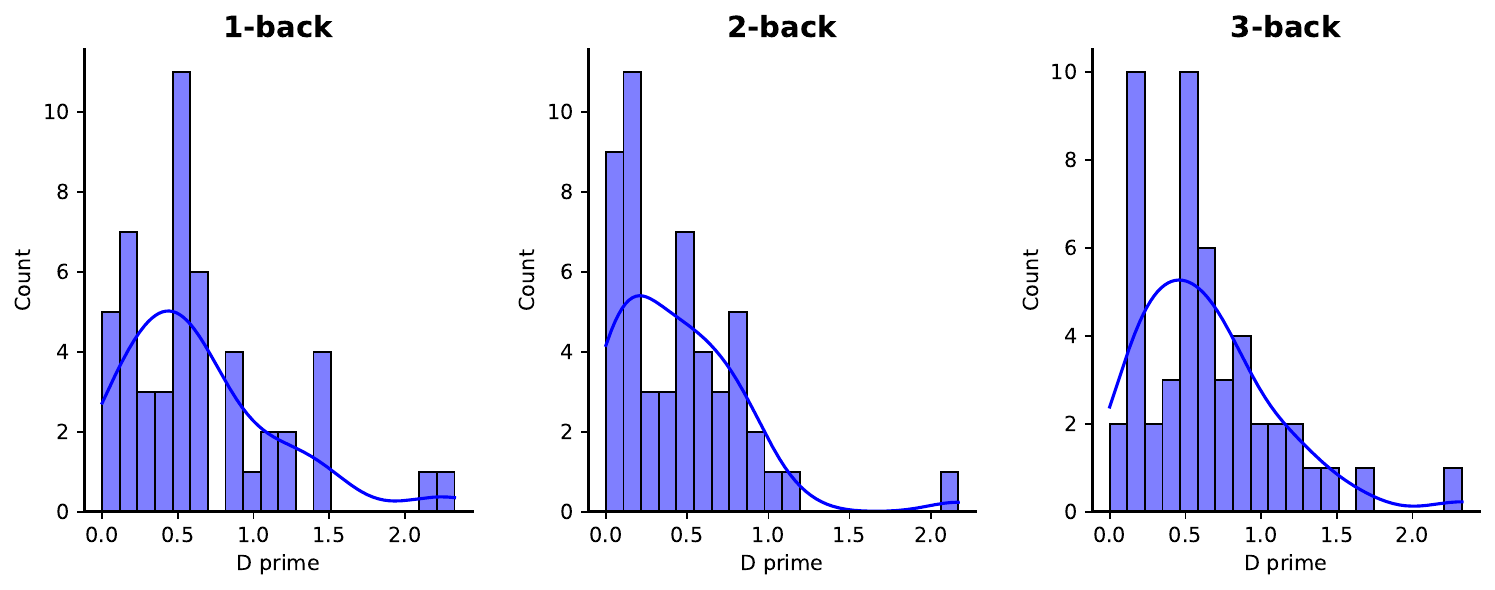} 
\caption{$d'$ distributions: spatial with abstract reasoning (excluding identical locations).}
\end{figure}


\begin{figure}[htb!]
\centering
\includegraphics[width=0.5\textwidth]{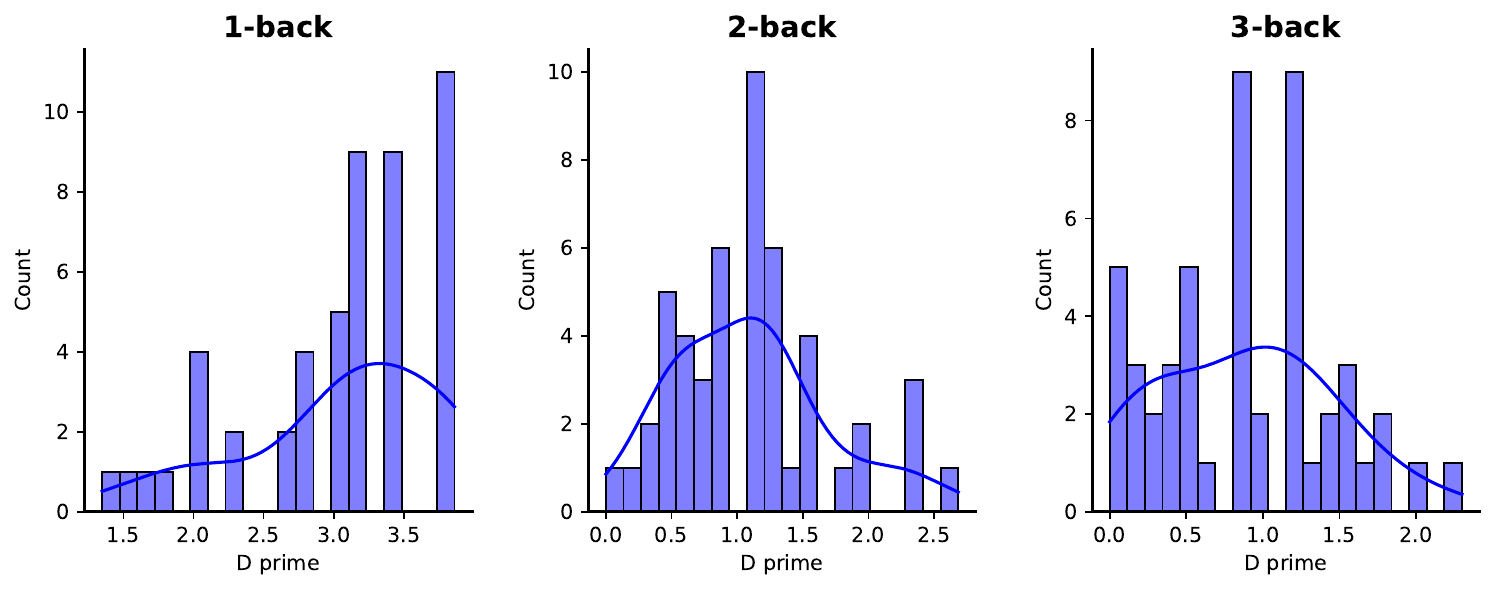} 
\caption{$d'$ distributions: spatial with 4*4 grids.}
\end{figure}


\begin{figure}[htb!]
\centering
\includegraphics[width=0.5\textwidth]{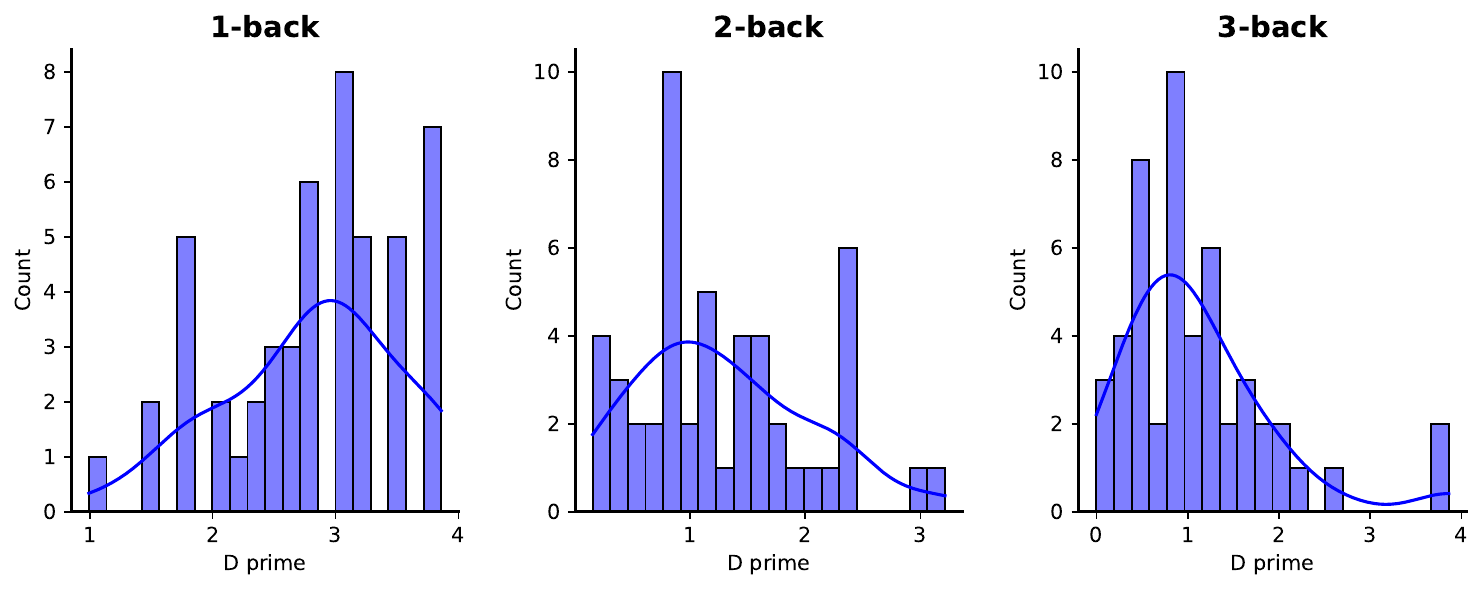} 
\caption{$d'$ distributions: spatial with 5*5 grids.}
\end{figure}


\begin{figure}[htb!]
\centering
\includegraphics[width=0.5\textwidth]{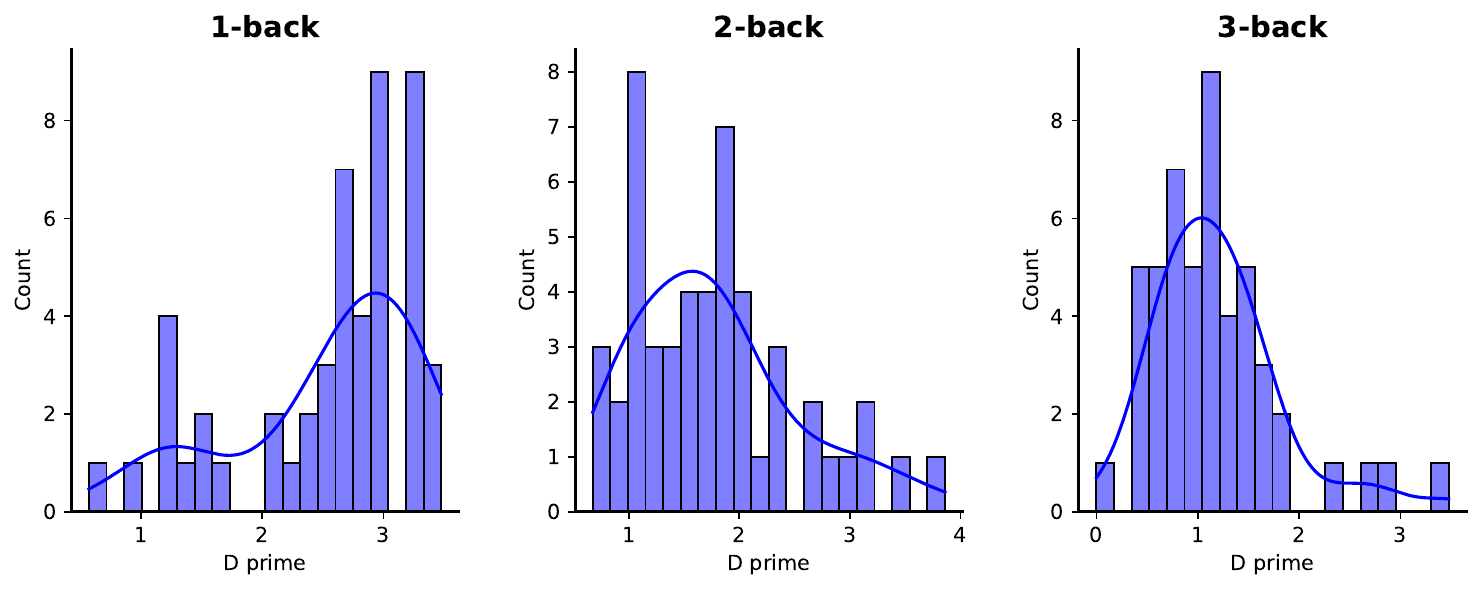} 
\caption{$d'$ distributions: spatial with 7*7 grids.}
\end{figure}

\begin{figure}[htb!]
\centering
\includegraphics[width=0.5\textwidth]{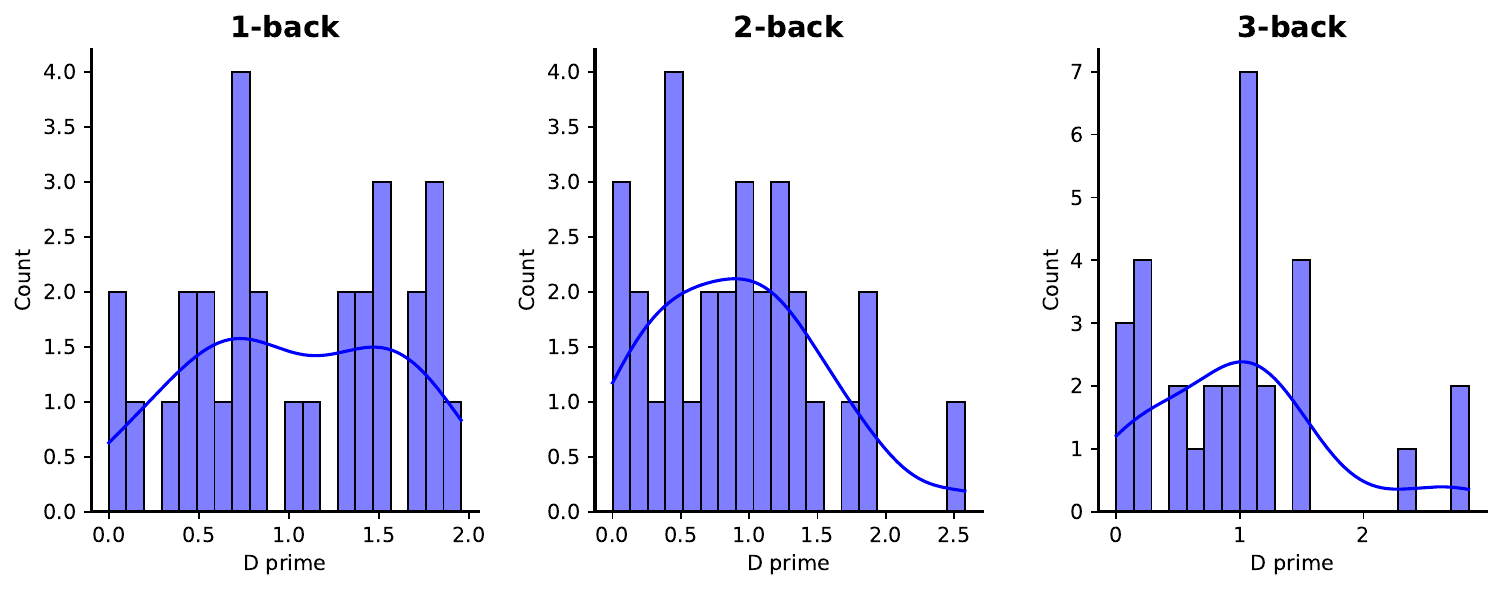}
\caption{$d'$ distributions: verbal (base version), using the Bloomz-7B model.}
\end{figure}

\begin{figure}[htb!]
\centering
\includegraphics[width=0.5\textwidth]{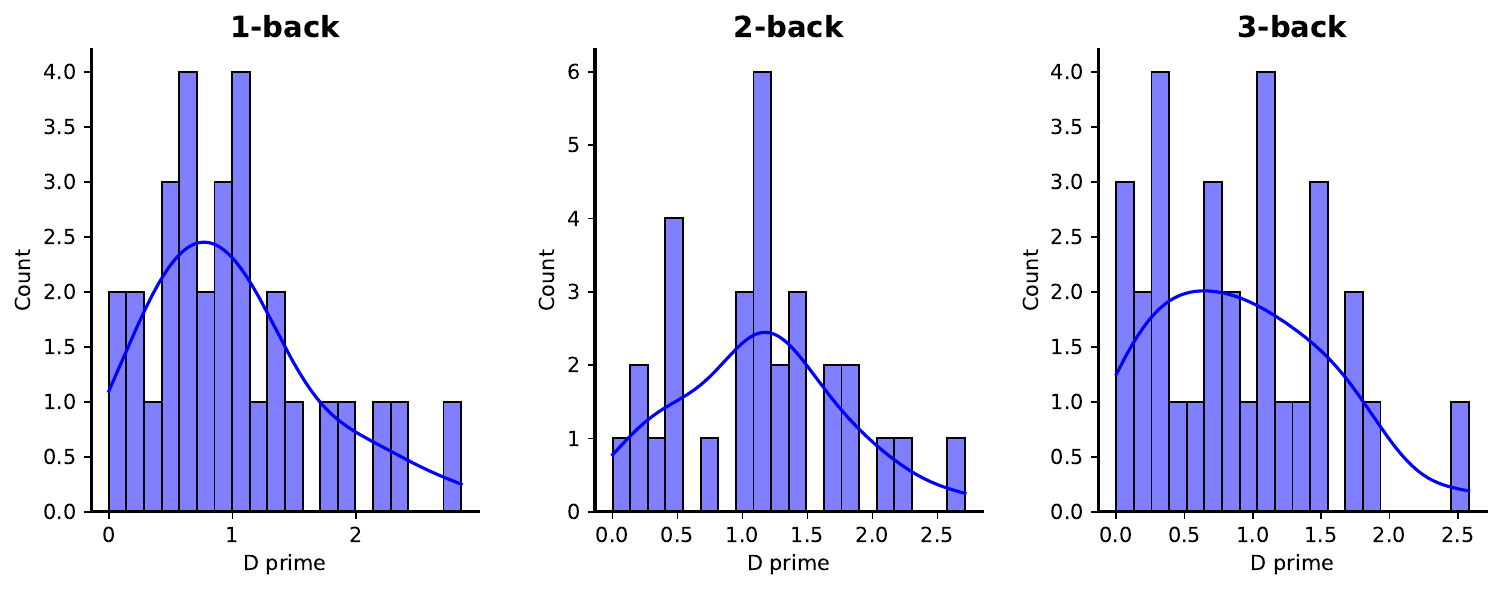}
\caption{$d'$ distributions: verbal (base version), using the Bloomz-7B1-mt model.}
\end{figure}

\begin{figure}[htb!]
\centering
\includegraphics[width=0.5\textwidth]{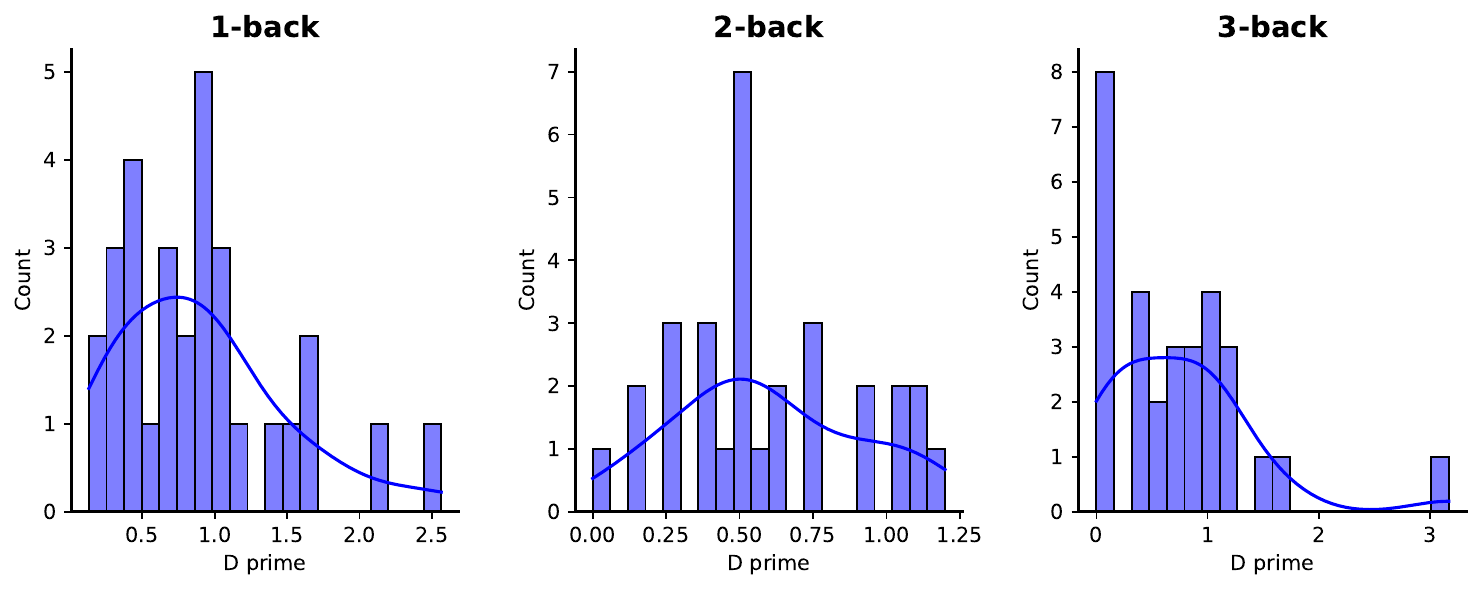}
\caption{$d'$ distributions: verbal (base version), using the ChatGLM-6B\_v1.0 model.}
\end{figure}

\begin{figure}[htb!]
\centering
\includegraphics[width=0.5\textwidth]{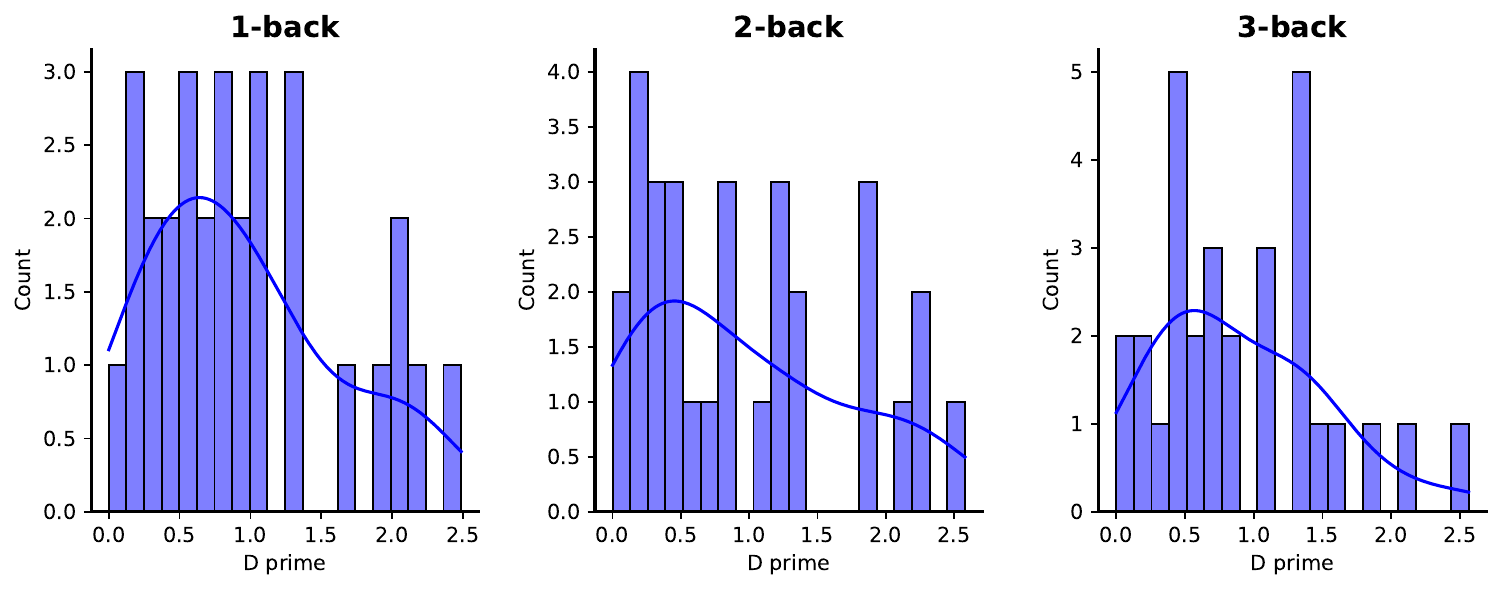}
\caption{$d'$ distributions: verbal (base version), using the ChatGLM-6B\_v1.1 model.}
\end{figure}

\begin{figure}[htb!]
\centering
\includegraphics[width=0.5\textwidth]{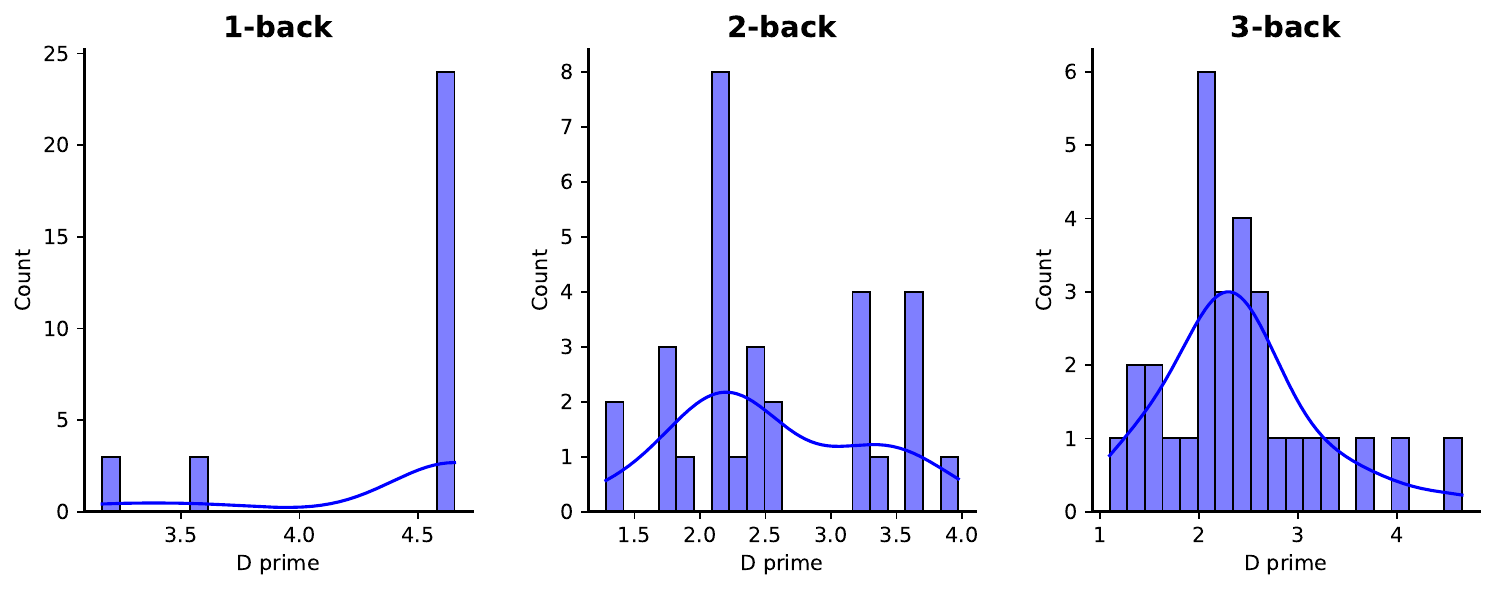}
\caption{$d'$ distributions: verbal (base version), using the GPT-4 model.}
\end{figure}

\begin{figure}[htb!]
\centering
\includegraphics[width=0.5\textwidth]{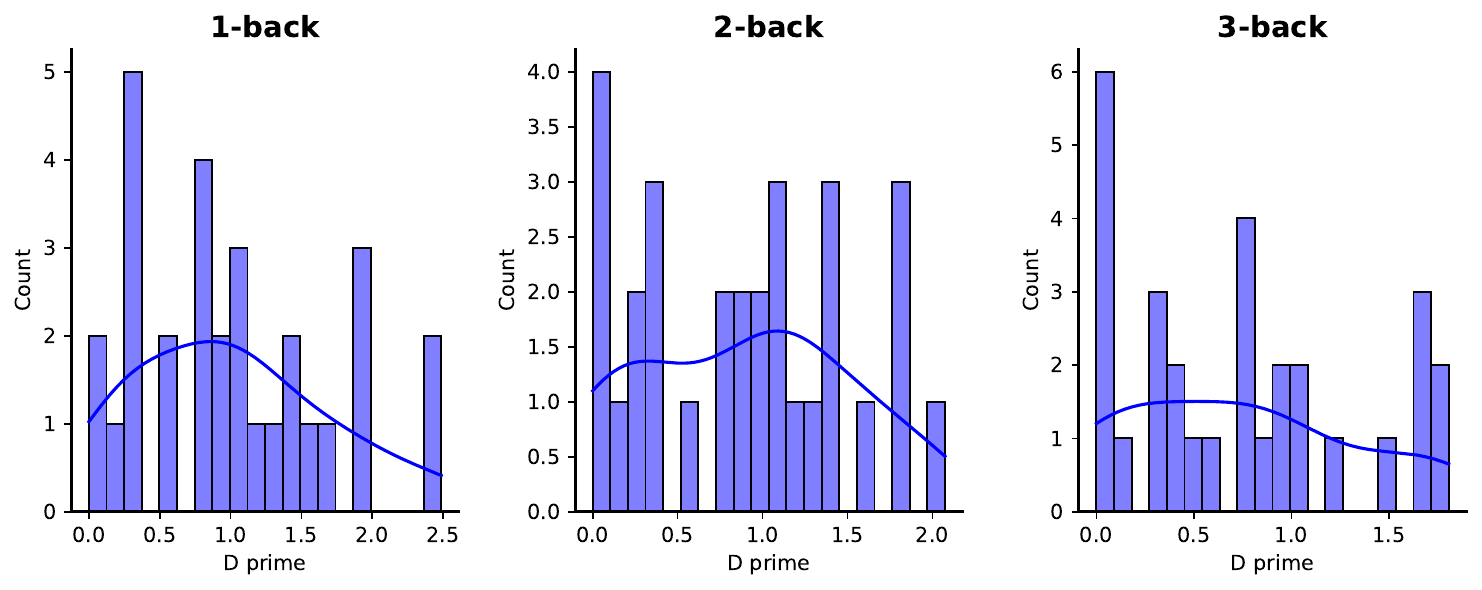}
\caption{$d'$ distributions: verbal (base version), using the Vicuna-7B model.}
\end{figure}

\clearpage

\begin{figure}[htb!]
\centering
\includegraphics[width=0.5\textwidth]{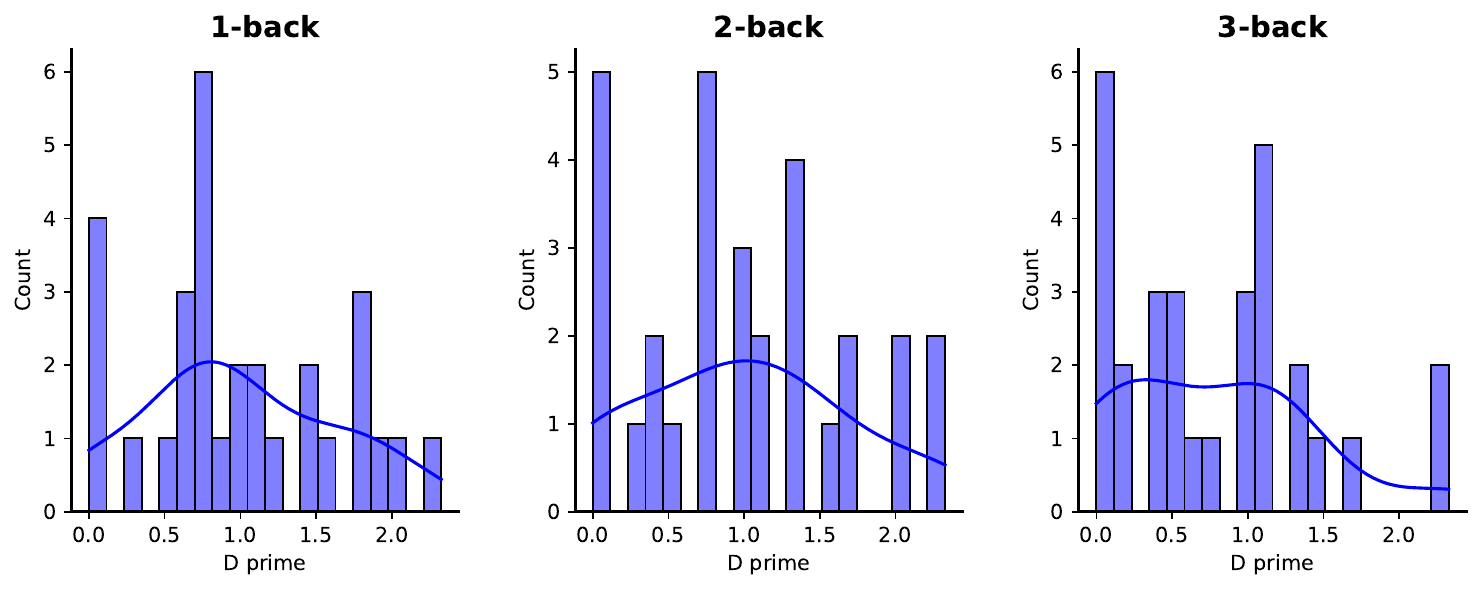}
\caption{$d'$ distributions: verbal (base version), using the Vicuna-13B model.}
\end{figure}